\documentclass[preprint,authoryear]{elsarticle}

\usepackage[T1]{fontenc}
\usepackage[utf8]{inputenc}
\usepackage{amsmath}
\usepackage{amsfonts}
\usepackage{amsthm}
\usepackage{enumitem} 
\usepackage[unicode=true, bookmarks=false, breaklinks=false,pdfborder={0 0 1},backref=false,colorlinks=false]{hyperref}

\usepackage{xcolor}
\usepackage{graphicx}
\usepackage{algorithm}
\usepackage{algorithmic}
\usepackage{booktabs} 
\usepackage{multirow}

\journal{Reliability Engineering \& System Safety}

\begin{document}

\begin{frontmatter}

\title{Algorithm-Informed Graph Neural Networks for Leakage Detection and Localization in Water Distribution Networks}

\author[first]{Zepeng Zhang}
\author[first]{Olga Fink}
\affiliation[first]{organization={Laboratory of Intelligent Maintenance and Operations Systems},
            addressline={EPFL}, 
            city={Lausanne},
            postcode={1015}, 
            country={Switzerland}}
\begin{abstract}
Detecting and localizing leakages is a significant challenge for the efficient and sustainable management of water distribution networks (WDN).
Given the extensive number of pipes and junctions in real-world WDNs, full observation of the network with sensors is infeasible.
Consequently, models must detect and localize leakages with limited sensor coverage.
Leveraging the inherent graph structure of WDNs, recent approaches have used graph interpolation-based data-driven methods and graph neural network models (GNNs) for leakage detection and localization.
However, these methods have a major limitation: data-driven methods often learn shortcuts that work well with in-distribution data but fail to generalize to out-of-distribution data.
To address this limitation and inspired by the perfect generalization ability of classical algorithms, we propose an algorithm-informed graph neural network (AIGNN) for leakage detection and localization in WDNs.
Recognizing that WDNs function as flow networks, incorporating max-flow information can be beneficial for inferring pressures.
In the proposed framework, we first train AIGNN to emulate the Ford-Fulkerson algorithm, which is designed for solving max-flow problems. This algorithmic knowledge is then transferred to address the pressure estimation problem in WDNs.
Specifically, two AIGNNs are deployed, one to reconstruct pressure based on the current measurements, and another to predict pressure based on previous measurements.
Leakages are detected and localized by analyzing the discrepancies between the outputs of the reconstructor and the predictor.
By pretraining  AIGNNs to reason like algorithms, they are expected to extract more task-relevant and generalizable features.
To the best of our knowledge, this is the first work that applies algorithmic reasoning to engineering applications.
Experimental results demonstrate that the proposed algorithm-informed approach achieves superior results with better generalization ability compared to GNNs that do not incorporate algorithmic knowledge.
\end{abstract}

\begin{keyword}
Graph neural networks \sep neural algorithmic reasoning \sep leakage detection and localization \sep water distribution networks



\end{keyword}

\end{frontmatter}
\section{Introduction}
Leakage is a persistent and unavoidable challenge faced by modern water distribution networks (WDNs). 
Despite advances in technology and improvements in infrastructure, the complex web of pipes and junctions that make up these systems is subject to various forms of wear and degradation. 
Factors such as corrosion, fluctuating pressures, environmental stresses, and material fatigue can lead to the development of leaks. 
Consequently, water systems around the world experience significant water loss, necessitating continuous monitoring, maintenance, and repair efforts to mitigate the impact and conserve as much water as possible \citep{puust2010review}.
The issue of water loss presents a substantial economic burden, with the estimated worldwide annual financial impact being around USD 39 billion in 2016 \citep{liemberger2019quantifying}. 
In addition to the economic problems for water utilities, the water loss issue also raises environmental and sustainability concerns \citep{puust2010review,gupta2017leakage}. 
To mitigate the consequences of such failures in WDNs, effective leakage detection and localization are of vital importance \citep{islam2022review,wan2022literature,romero2023leak}. 

Existing approaches for leakage detection and localization can be categorized into model-based and data-driven methods \citep{chan2018review,romero2023leak}. 
Model-based approaches involve constructing a hydraulic model to simulate the behavior of WDNs \citep{sanz2016leak,sophocleous2019leak,steffelbauer2022pressure}. 
These hydraulic models typically characterize the state evolution of WDNs using physics-based equations, providing high-fidelity predictions of WDN behavior.
Leakages are detected by comparing collected real-world hydraulic data to the simulated data.
Ideally, a well-calibrated hydraulic model can facilitate optimal detection.
However, constructing and maintaining such an accurate model is challenging in practice due to several practical obstacles. These include (1) uncertainties regarding model parameters, such as the varying condition of pipes, which can make  models based on static pipe conditions inaccurate;
(2) incomplete  documentation of WDNs' system information  due to the age and heterogeneity of these networks; and
(3) unavailability of real-time consumer demand information, which is crucial for hydraulic models \citep{cuguero2016model}.

As physics-based hydraulic models are not always available or adequately calibrated, an increasing number of data-driven approaches have recently been developed \citep{soldevila2020leak,romero2022leak,gardharsson2022graph}.
Data-driven approaches rely solely on data collected by monitoring devices and do not depend on in-depth knowledge of the system, such as the condition of the pipes or consumers' demand.
Typically,  data-driven models are trained using historical measurements collected from monitoring devices within the network under normal operating conditions.
Leakages are then detected by measuring the deviations between real-world hydraulic data and predictions generated by data-driven models \citep{romero2023leak}.
Previous approaches have employed various techniques such as support vector machines \citep{mounce2011novelty}, evolutionary polynomial regression \citep{laucelli2016detecting}, and convolutional neural networks \citep{fang2019detection}.

In practice, it is too expensive to monitor every junction in a WDN, resulting in limited sensor coverage.
Given that WDNs are inherently represented as graphs with junctions as nodes and pipes as edges, graph-based interpolation methods have been widely used to handle partially observed WDNs \citep{romero2023leak}.
Such methods typically involve two steps: first, training a data-driven model to estimate the expected leak-free pressure at monitored nodes based on historical measurements, and then performing a graph-based interpolation to approximate the expected leak-free pressure at unmonitored nodes \citep{soldevila2020leak,romero2022leak}.
One limitation of these graph-based interpolation methods is that they are post-processing techniques, making the data-driven models agnostic to the graph topology.
A more holistic design that integrates the interpolation process within the data-driven model could potentially yield further improvements.
Recently, \citet{gardharsson2022graph} used graph neural networks (GNNs) as the data-driven model, which can perform pressure estimation and interpolation simultaneously.
Specifically, this approach employs two separate GNNs: a prediction network to predict the leak-free pressures of the entire  WDN based on historical measurements, and a reconstruction network to reconstruct the actual pressures of the whole WDN based on current partially observed measurements.
Leakages are detected and localized by analyzing the differences between the outputs of these two networks.

A major limitation of data-driven approaches is that deep learning models often learn shortcuts that work well with in-distribution data but fail to generalize to out-of-distribution data \citep{wang2022generalizing}.
This generalization ability is especially important for handling WDNs, as deep learning models are often trained based on simulated data generated by mathematical hydraulic models.
Given the inherent uncertainty of real-world scenarios, data-driven models are likely to encounter distribution shifts.
For instance, changes in pipe conditions or network structures can alter the data distribution,  posing significant challenges to model accuracy and reliability.
In contrast, classical algorithms are renowned for their perfect generalization ability and interpretability \citep{cormen2022introduction}.
For example, the Bellman-Ford algorithm for shortest-path finding will always give the optimal solution regardless of the data distribution \citep{cormen2022introduction}.
However, they require input data in a strict format, which can make them challenging to apply to real-world problems.
For example, the Bellman-Ford algorithm can only handle graphs with scalar edge weights, whereas real-world shortest-path finding problems are often too complex to be accurately represented by simple scalar values.
To leverage the strengths of both deep learning models and classical algorithms, recent advancements in neural algorithmic reasoning (NAR) aim to construct neural networks capable of approximating and executing classical algorithms \citep{velivckovic2021neural,velivckovic2022clrs,cappart2023combinatorial}.
With the ability to reason like algorithms, NAR models are expected to generalize well to out-of-distribution data\citep{mahdavi2023towards,rodionov2024discrete,de2024simulation}.
Typically, NAR models employ an encoder-processor-decoder framework. The encoder and decoder transform the data into a latent space that the processor can operate on, while the processor learns the step-by-step execution of the target algorithms \citep{velivckovic2022clrs}. 
 Once the processor has acquired the algorithmic knowledge, it can be reused to solve real-world problems where such algorithmic knowledge is beneficial \citep{velivckovic2021neural}.
For example, \citet{georgiev2023narti} transfer the algorithmic knowledge of Prim's minimum spanning tree algorithm to reconstruct the developmental trajectory of single cells from high-dimensional gene expression data. Similarly,\citet{numeroso2023dual} transfer the algorithmic knowledge of max-flow and min-cut algorithms to help classify the types of vessels in brain graphs.

In this paper, inspired by the achievements of NARs \citep{velivckovic2022clrs}, we propose leveraging algorithmic knowledge to enhance the detection and localization of leakages in WDNs.
Specifically, because  WDNs function as flow networks, incorporating max-flow information helps in accurately inferring pressures within the network.
We propose a method to transfer the knowledge of the Ford-Fulkerson algorithm, which solves max-flow problems, to  GNN models for pressure inference in WDNs.
Consequently, we name our model algorithm-informed GNN (AIGNN). 
Following the paradigm outlined in \citep{gardharsson2022graph}, we employ a prediction AIGNN to predict the leak-free pressures and a reconstruction AIGNN to reconstruct actual pressures within the WDN.
Leakages are then detected and localized by analyzing the deviations between the reconstructed and predicted pressures.
Experimental results demonstrate that the AIGNN outperforms the ChebNet model proposed in \citep{gardharsson2022graph}, which is a GNN model without algorithmic knowledge.
Additionally, AIGNN exhibits superior generalization ability compared to ChebNet.
Furthermore, integrating the embeddings generated by AIGNN improves the performance of ChebNet, indicating that AIGNN can serve as an auxiliary enhancement for existing models.
The main contributions of this paper are summarized as follows:
  
\begin{enumerate}[label=(\arabic*)]
  \item We propose an algorithm-informed data-driven model for leakage detection and localization in WDNs, named AIGNN. 
  AIGNN is trained to reason following the principles of the Ford-Fulkerson algorithm for solving max-flow problems. 
  To the best of our knowledge, this is the first application of NAR in the context of engineering applications. 
  
  \item The proposed AIGNN extracts algorithm-informed information from the network, significantly enhancing its generalizability. 
  Building on existing GNN-based approaches, we train two AIGNNs to approximate the actual pressure and another to estimate the leak-free pressure of the entire  WDN. 
  Leakages are then detected and localized by comparing the differences between the estimated actual and leak-free pressures.
  
  \item The embeddings generated by AIGNN can also be used as augmentations for other methods, providing additional information extracted from network features.
  We integrate the embeddings of AIGNN with either the input layer or the embedding of the layer just before the final layer of the ChebNet model, resulting in the variants ChebNet$_\text{IN}$ and ChebNet$_\text{EMB}$, respectively.
  \item We evaluate the proposed AIGNN on the $L$-town WDN.
  The AIGNN model outperforms the ChebNet model in both pressure prediction and reconstruction.
  Additionally, augmenting the baseline ChebNet model with AIGNN embeddings further improves its performance.
  Algorithm-informed methods also exhibit superior results in leakage detection and localization.
  Experiments further confirm the superior generalization capability of AIGNN, demonstrating that it outperforms ChebNet when trained on one set of sensor locations and tested with sensors placed at different locations.
\end{enumerate}

The remaining content of this paper is structured as follows.
In Section \ref{sec:related_work}, we review related work on pressure estimation in WDN and NAR.
 Section \ref{sec:preliminaries} provides the problem definition for leakage detection and localization in WDNs, as well as an overview of the basic architecture of NAR models.
Section \ref{sec:proposed_method} presents our proposed method, AIGNN.
In Section \ref{sec:experiments}, we evaluate the effectiveness and the generalization ability of AIGNN through a case study using the $L$-town dataset.
Finally, in Section \ref{sec:conclusions}, we summarize the main conclusions of this work.

\section{Related Work}\label{sec:related_work}
\subsection{Pressure Estimation in Water Distribution Networks}
In WDNs, leakages can be detected by analyzing the system's pressure or flow \citep{puust2010review}.
Pressure measurements are more commonly used for leakage detection and localization because pressure sensors are easier to install and less costly compared to flow sensors \citep{zhou2019deep}.
Large and widespread WDNs can contain thousands of junctions, making it infeasible to employ and maintain sensors at every junction due to infrastructural limits, privacy concerns, and high costs \citep{truong2023graph}.
Consequently, estimating the complete network pressure based on partially observable pressure measurements is typically the first and essential step for leakage detection and localization.

There are two main types of approaches for pressure estimation in WDNs: model-based and data-driven methods.
Model-based methods rely on mathematical hydraulic simulation tools, which can deliver optimal results when well-calibrated.
However, due to uncertainties in real-world scenarios and incomplete documentation of WDNs' system information, achieving accurate hydraulic simulation is often challenging.
Data-driven approaches often use deep learning models to estimate complete pressure profiles from partial nodal pressure data\citep{hajgato2021reconstructing,xing2022graph,ashraf2023spatial}.
During inference, these models only require data collected by monitoring devices, eliminating the need for in-depth system knowledge.
Since WDNs can inherently be modeled as graphs, graph-based approaches can leverage the relational inductive biases to enhance model performance  \citep{wu2021comprehensive}.
The graph interpolation methods typically involve two steps: first, training a data-driven model to predict pressures at monitored nodes, and then using graph interpolation techniques to infer pressures at unmonitored nodes \citep{soldevila2020leak,romero2022leak}.
Specifically, \citet{soldevila2020leak} use the Kriging interpolation method, while  \citet{romero2022leak} address the interpolation by solving a quadratic optimization problem to estimate pressure signals.
However, these approaches have limitations. The deep learning models used are typically unaware of the graph topology, and the subsequent graph interpolation process implicitly assumes specific smoothness patterns of the node features, often disregarding the actual measurements at monitored nodes \citep{soldevila2020leak,romero2022leak}.

\citet{hajgato2021reconstructing} took the first step in addressing the challenges encountered by the graph-based interpolation methods by using a Chebyshev spectral convolutional neural network (ChebNet) to estimate pressures. ChebNet, a topology-aware deep learning model, can also perform interpolation. The ChebNet model has demonstrated superior performance compared to traditional graph-based interpolation methods \citep{hajgato2021reconstructing}.
Building on this, \citet{xing2022graph} developed a physics-informed GNN that incorporates the physical functional relationships between monitored and unmonitored locations as an additional loss during the training process.
However, this approach requires demand patterns from every consumer, even during inference time.
Beyond pressure estimation, GNNs have also demonstrated strong performance in other state estimation tasks in WDNs, such as demand prediction \citep{zanfei2022graph} and water quality prediction \citep{li2024real}.

\subsection{Neural Algorithmic Reasoning}
NAR is an emerging research area focused on developing neural networks that can perform  algorithmic computations within  a high-dimensional latent space \citep{velivckovic2021neural, velivckovic2022clrs, minder2023salsa}.
Algorithms inherently  provide  correct solutions with theoretical guarantees.
For instance, the Bellman-Ford algorithm for shortest-path finding consistently  generates the optimal solution, irrespective  of the graph's properties \citep{cormen2022introduction}.
However, applying algorithms to real-world problems requires translating real-world scenarios into the abstract spaces in which algorithms operate, a longstanding and  significantly challenging issue \citep{harris1955fundamentals}.
Furthermore, classical algorithms often  become impractical  when dealing with partially observable data.
Thus, despite their theoretical  correctness, the outputs of classical algorithms are  approximations, because the abstract space representation may not precisely describe the real-world problem of interest.

Instead of manually translating real-world scenarios into the abstract spaces that algorithms operate in, NAR promotes algorithm execution using neural networks, known for their high flexibility in handling different types of inputs, making this approach more adaptable to real-world complexities.
Typically, the NAR model employs an encoder-processor-decoder framework \citep{velivckovic2021neural}. 
The encoder projects real-world inputs into the abstract space, while the decoder maps the embeddings generated by the processor back to the real-world output space. 
In classical algorithms such as Bellman-Ford for shortest path finding \citep{ford1956network,bellman1958routing}, solutions are typically obtained through iterative steps, producing an output along with a trajectory of intermediate steps.
To align the processor with the underlying classical algorithm, it is usually designed based on some form of recurrent unit, aiming to replicate the full trajectory of algorithmic behaviors.
Given the theoretical evidence supporting that GNN models align well with classical algorithms, GNNs are the most commonly used processors \citep{xu2019can,xu2020neural,dudzik2022graph}.
The recently proposed CLRS-30 algorithmic reasoning benchmark \citep{velivckovic2022clrs} provides a general paradigm of algorithmic reasoning, named in homage to the \textit{Introduction to Algorithms} textbook by Corman, Leiserson, Rivest, and Stein \citep{cormen2022introduction}. 
The CLRS-30 benchmark covers 30 algorithms spanning sorting, searching, dynamic programming, geometry, graphs, and strings,  all uniformly represented over the graph domain.
To provide supervision on the algorithmic trajectories, CLRS-30 decomposes these trajectories into consecutive execution steps, called hints \citep{velivckovic2022clrs}.
With the supervision of hints, the NAR model is expected to align better with the underlying algorithm and have fewer dependencies on the non-generalizable features of a particular training set.
In addition to learning individual algorithms,  some works investigate the possibility of tackling multiple algorithmic tasks with a single neural reasoner \citep{xhonneux2021transfer,ibarz2022generalist}.

Despite the advancements made by recent NAR models in the CLRS-30 benchmark, most studies focus primarily on learning algorithms themselves  \citep{minder2023salsa,mahdavi2023towards,Montgomery2024markov}.
However, the transfer of algorithmic knowledge to solve real-world problems -- advocated as the major motivation for designing NAR models \citep{velivckovic2021neural} -- remains under-explored. 
To date,  only a few works have explored the use of NAR in biology and computer network configuration problems \citep{beurer2022learning,georgiev2023narti,numeroso2023dual}. 
The application of NAR to other real-world problems, especially in engineering, remains largely unexplored. 

\section{Preliminaries}\label{sec:preliminaries}
\subsection{Problem Statement}
We address the problem of leakage detection and localization in a partially observable WDN, where sensors are installed only at select junctions.
The WDNs are inherently structured as graphs, with edges representing pipes and nodes representing junctions.
Specifically, the topological structure of a WDN can be represented by a weighted undirected graph $\mathcal{G}=(\mathcal{V},\mathcal{E}, \mathcal{W})$, where $\mathcal{V}$, $\mathcal{E}$, and $\mathcal{W}$ denote the sets of nodes, edges, and weights, respectively. 
The graph adjacency matrix is given by $\mathbf{A}\in\mathbb{R}^{N\times N}$, where $N$ is the number of nodes.
We denote by $\mathbf{1}$ the all-one vector and by $\mathbf{I}$ the identity matrix. 
Given $\mathbf{D}=\mathrm{Diag}\left(\mathbf{A}\mathbf{1}\right)\in\mathbb{R}^{N\times N}$ as the diagonal degree matrix, the Laplacian matrix is defined as $\mathbf{L}=\mathbf{D}-\mathbf{A}$. 
The symmetric normalized adjacency matrix is then defined as $\mathbf{A}_{{\rm sym}}=\mathbf{D}^{-\frac{1}{2}}\mathbf{A}\mathbf{D}^{-\frac{1}{2}}$, which is commonly used in GNN models instead of the standard adjacency matrix.
For notational simplicity, we will use $\mathbf{A}$ to represent the symmetric normalized adjacency matrix in the following sections.
We denote by $\mathbf{X}\in\mathbb{R}^{N\times M}$ the node feature matrix, where $M$ is the number of dimensions of the node features, and its $i$-th row $\mathbf{X}_{i,:}$ represents the feature vector corresponding to the $i$-th node, $i=1,\ldots,N$. 
In this work, following the setting of previous studies \citep{hajgato2021reconstructing,gardharsson2022graph,truong2023graph}, the node features include not only the nodal pressures but also an additional binary feature to indicate the locations of the sensors.
Given that the network is partially observable, the feature matrix $\mathbf{X}$ is sparse, with nonzero elements only in the rows corresponding to nodes equipped with sensors.
At each timestep $t$, the goal is to perform leakage detection and localization based on the current and previous pressure data collected from the partially deployed sensors, denoted as $\mathbf{X}_{\mathcal{G}}(t),\mathbf{X}_{\mathcal{G}}(t-1),\ldots,\mathbf{X}_{\mathcal{G}}(t-T)$.

\subsection{Neural Algorithmic Reasoning}\label{sec:NAR}
\begin{figure*}
    \centering
    \includegraphics[width=1.0\textwidth]{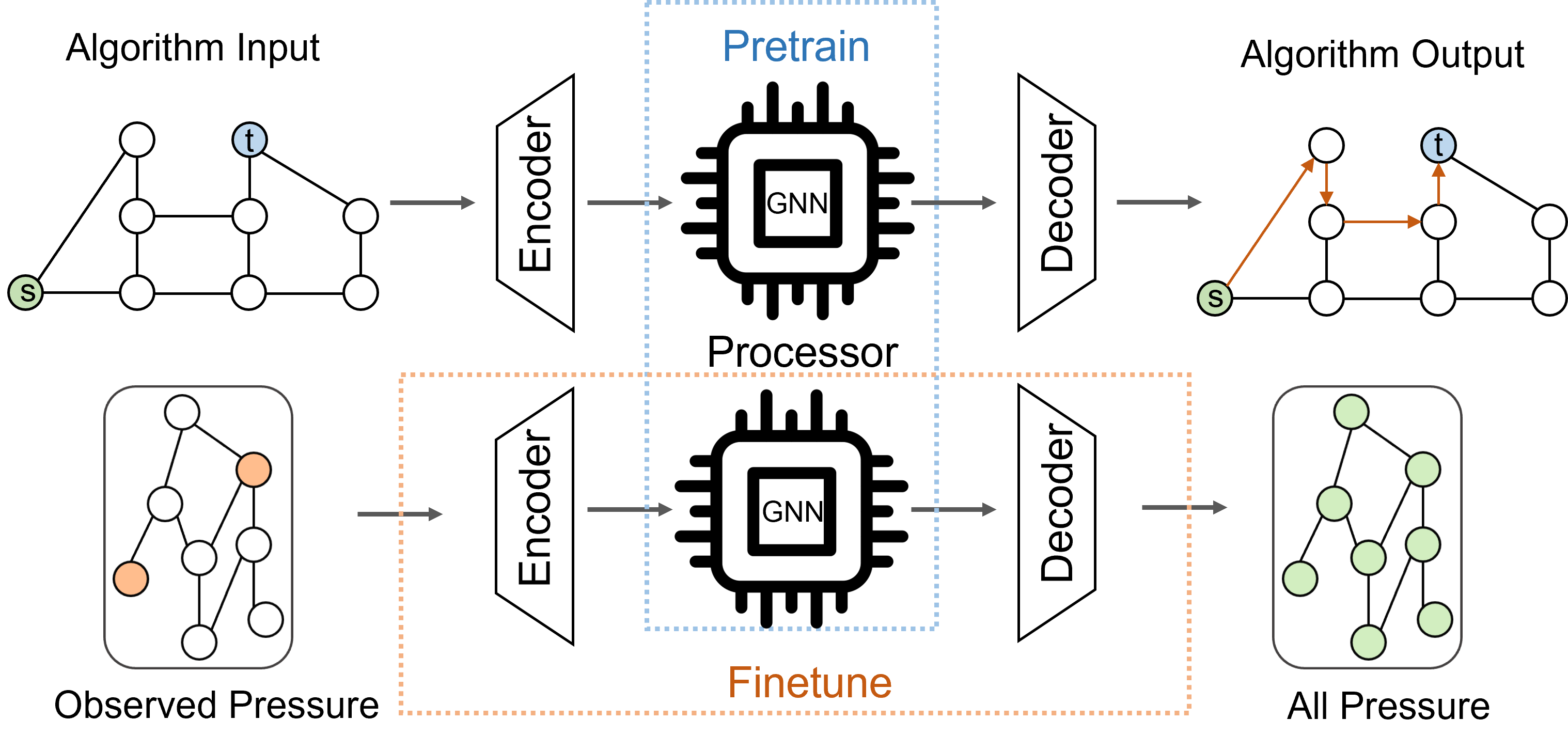}
    \caption{The architecture of the algorithm-informed reconstructor and predictor.\label{fig:NAR}}
\end{figure*}

Our work follows the encoder-processor-decoder paradigm \citep{hamrick2018relational}, a standard approach for step-by-step neural execution, particularly in the field of NAR \citep{velivckovic2021neural,cappart2023combinatorial}.
Consequently, a NAR model comprises three components: the encoder $f$, the processor $p$, and the decoder $g$, as depicted in the upper half of Figure \ref{fig:NAR}.
We consider the algorithmic reasoning tasks as formulated in the CLRS-30 benchmark \citep{velivckovic2022clrs}.
In the CLRS-30 benchmark, the intermediate states of the algorithm are captured by hints, which describe the characteristics that can determine the steps of the algorithm.
Specifically, for a target algorithm, each data sample contains the execution trajectory of the algorithm, represented by the inputs, outputs, and hints.

Let us denote the node-level and edge-level features at algorithmic step $k$ as $\{\mathbf{x}_i^{(k)}\}$ and $\{\mathbf{e}_{ij}^{(k)}\}$, $k=1,\ldots,K$. 
Here, $i$ indicates the node index and $ij$ specifies the index of the edge connecting node $i$ and node $j$.
While it is also possible to consider problems with graph-level features, the leakage detection and localization problem addressed in this paper does not involve such features, so we omit them for simplicity.
At algorithmic step $k$, the encoder first embeds the current features into high-dimensional representations as follows:
\[
\begin{aligned}\mathbf{z}_{i}^{(k)} & =f_{n}\left(\mathbf{x}_{i}^{(k)}\right),\\
\mathbf{d}_{ij}^{(k)} & =f_{e}\left(\mathbf{e}_{ij}^{(k)}\right),
\end{aligned}
\]
where $f_n(\cdot)$ and $f_e(\cdot)$ represent encoder layers that encode the node-level and edge-level features, respectively. 
The encoded embeddings are then fed into a processor.
The processor in NAR is considered the central component, as it is expected to learn to reason like algorithms.  
Motivated by theoretical findings that GNNs align well with classical algorithms \citep{xu2019can,xu2020neural,dudzik2022graph}, existing NAR models normally employ GNNs as processors.
Thus, we denote the processor by $p_{\mathsf{GNN}}(\cdot)$. 
Depending on the computational dynamics of the target algorithms, the specific GNN architecture is chosen to better align with the target algorithms \citep{velivckovic2020pointer,numeroso2023dual}.
In general, the processor can be formulated as:
\[
\begin{aligned}\{\mathbf{h}_{i}^{(k)}\} & =p_{\mathsf{GNN}}\left(\mathbf{z}_{i}^{(k)},\mathbf{h}_{i}^{(k-1)},\{\mathbf{z}_{j}^{(k)}\}_{j\in\mathcal{N}(i)},\{\mathbf{h}_{j}^{(k-1)}\}_{j\in\mathcal{N}(i)},\{\mathbf{d}_{ij}^{(k)}\}_{(i,j)\in\mathcal{E}}\right),\end{aligned}
\]
where $\mathbf{h}_{i}^{(k)}$ represents the embedding vector of node $i$ at the $k$-th layer of the processor and $\mathcal{N}(i)$ is the set of all neighboring nodes of node $i$. 
The initial embedding vector $\mathbf{h}_{i}^{(0)}$ is initialized as $\mathbf{h}_{i}^{(0)}=\mathbf{0}$ for all nodes.
The processor recurrently processes the encoded features, similar to the classical algorithm.
Subsequently, the decoders map the embeddings produced by the processor to the output space of the algorithm:
\[
\begin{aligned}\mathbf{y}_{i}^{(k)} & =g_{n}\left(\mathbf{x}_{i}^{(k)}\right),\\
\mathbf{c}_{ij}^{(k)} & =g_{e}\left(\mathbf{e}_{ij}^{(k)}\right),
\end{aligned}
\]
with $g_n$ and $g_e$ representing decoder layers that decode the node-level and edge-level embeddings, respectively.
Depending on the stage of the process, the output of the decoder can be the hints for the next algorithmic step, or the final output of the algorithm if it is at the final algorithmic step.

To train the described NAR model, the loss is calculated based on the decoded hints at every step and the final output.
The hint supervision constrains the network to behave more closely to the target algorithm at each intermediate step $k$.
This ensures that the NAR model emulates the algorithmic process rather than merely learning the input-output mapping.
This approach enhances the interpretability of the model and improves its generalizability \citep{velivckovic2022clrs,bevilacqua2023neural,minder2023salsa}. 

\section{The Proposed Method}\label{sec:proposed_method}
We address the problem of leakage detection and localization in WDNs, where sensors are sparsely deployed in the network to collect pressure data. 
Given the partial availability of pressure data, an intuitive strategy involves first estimating both the actual pressure and the leak-free pressure of the complete WDN, and then identifying leakages by analyzing the differences between these estimated signals \citep{soldevila2020leak,romero2022leak,gardharsson2022graph}.
Specifically, we adopt the two-stage scheme used in \citep{gardharsson2022graph}. 
In the first stage, we construct an algorithm-informed reconstructor to estimate the actual nodal pressures for all nodes based on the currently observed data and an algorithm-informed predictor to predict the leak-free nodal pressures of all nodes based on previously collected measurements. 
In the second stage, we calculate the residual pressure by comparing the outputs of the reconstructor and the predictor. 
By analyzing these residuals, we can effectively detect and locate leakages within the WDN.

\begin{figure*}
    \centering
    \includegraphics[width=1.0\textwidth]{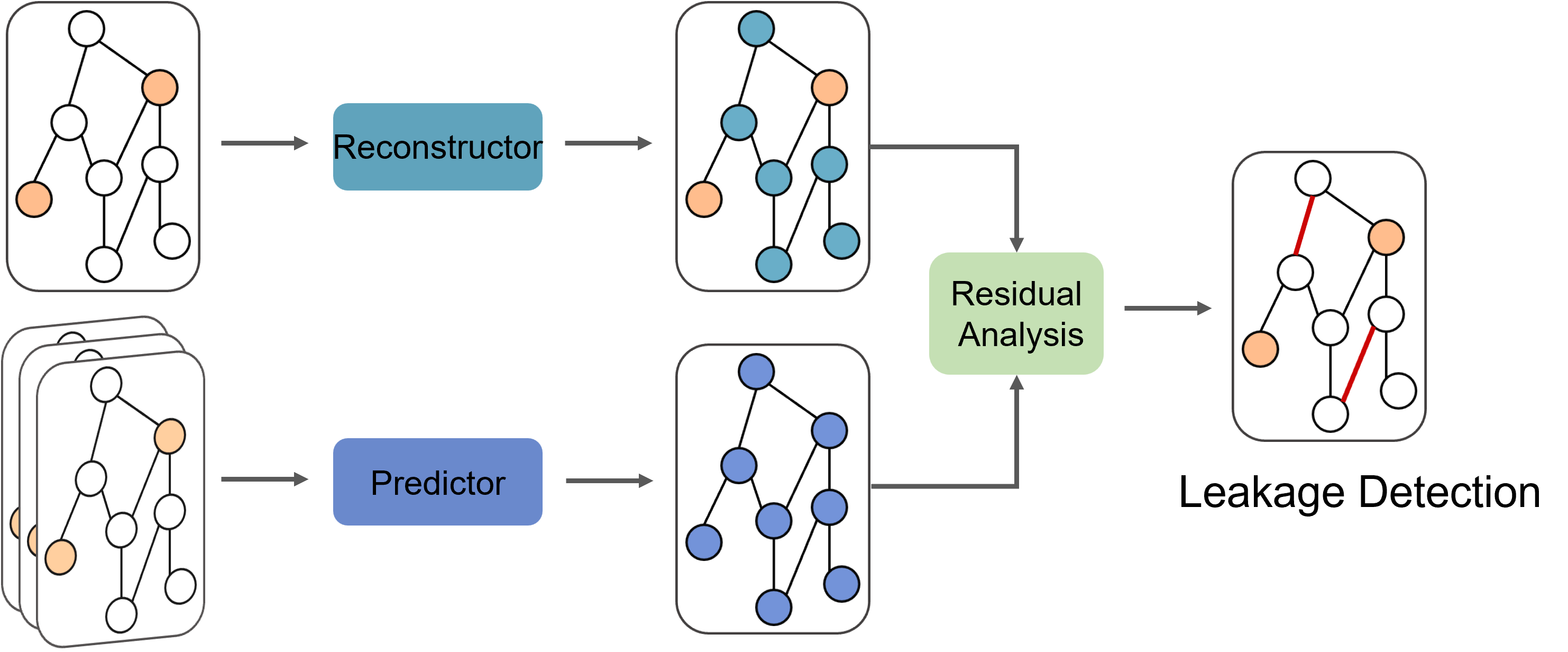}
    \caption{The pipeline of the proposed two-stage method.}
    \label{fig:overall_pipeline}
\end{figure*}

The overall pipeline of the approach is depicted in Figure \ref{fig:overall_pipeline}. In the following sections, we will detail each component of the model.

\subsection{First Stage: Algorithm-Informed Reconstructor and Predictor}
In the first stage, two AIGNNs are trained: 
(i) \textbf{Reconstructor:} This AIGNN takes $\mathbf{X}_{\mathcal{G}}(t)$ as the input and outputs the complete actual pressure signal $\hat{\mathbf{y}}_{r}(t)$; 
(ii) \textbf{Predictor:} This AIGNN takes the set of previous sparse graph pressure signals $\mathbf{X}_{\mathcal{G}}(t-T),\ldots,\mathbf{X}_{\mathcal{G}}(t-1)$ as the input and outputs the complete leak-free pressure signal $\hat{\mathbf{y}}_{p}(t)$.
The reconstructor and the predictor are trained using data simulated by mathematical hydraulic models under leak-free conditions. 
Intuitively, the reconstructor estimates the pressure data at all nodes based on the current conditions, while the predictor estimates what the pressure should be in the absence of leaks.
Therefore, when the output of the reconstructor significantly deviates from the output of the predictor, it is indicative of a potential leak.

Instead of directly training the model to learn the input-output mappings, we propose to enhance the model with algorithmic knowledge.
The flow properties of a water network are highly informative for inferring pressures, and the ability to simulate the algorithm for max-flow in the entire network is likely to be advantageous for this task. 
Motivated by this, we propose transferring the algorithmic knowledge of solving max-flow problems to assist in pressure reconstruction and prediction tasks in a WDN.

Consider a graph $\mathcal{G}=(\mathcal{V},\mathcal{E})$ where each edge  $\forall (u,v)\in\mathcal{E}$ has certain capacity $c_{uv}$. 
The max-flow from the source node $v_s\in\mathcal{V}$ to the sink node $v_t\in\mathcal{V}$ is the maximum flow that can pass through the graph without exceeding the capacities. 
To find the max-flow, we focus on the neural execution of the Ford-Fulkerson algorithm \citep{ford1956maximal}. 
We collect the edge capacities in a capacity matrix $\mathbf{C}$ and we construct a flow matrix $\mathbf{F}$ to represent the flow information of the entire network when the max-flow is achieved.
The Ford-Fulkerson algorithm has two key subroutines:
\begin{enumerate}
    \item Finding a valid path $\mathcal{P}=\{v_s, \ldots, v_t\}$ from the source node $v_s$ to the sink node $v_t$.
    \item Augmenting the residual capacity along the path based on the maximum flow $c_p$ that can pass through this path, then updating the capacity matrix $\mathbf{C}$ and the flow matrix $\mathbf{F}$.
\end{enumerate}
The output of the algorithm is the flow matrix $\mathbf{F}$.
We represent the max-flow solution with the flow matrix $\mathbf{F}$ instead of the commonly used max-flow value because the flow matrix characterizes the flow properties of the entire network. 
This is advantageous when addressing the pressure reconstruction and prediction problem in water networks.
The overall algorithm is outlined in Algorithm \ref{alg:ff}.

In the Ford-Fulkerson algorithm, the augmenting path can be obtained by various methods, such as breadth-first search and depth-first search.
However, since our goal is to train an NAR model to execute the algorithm, providing proper supervision based on these methods is challenging due to the potential existence of multiple valid paths for both breadth-first search and depth-first search.
As suggested in \citep{georgiev2020neural}, we use the Bellman-Ford algorithm to find the shortest path.   
Although, in theory,  multiple valid shortest paths can exist, this seldom happens in practice, and the effect can be considered negligible.
The Bellman-Ford algorithm is also appealing because it has been shown that GNNs align well with it \citep{xu2019can}.

\begin{algorithm}[ht]
\caption{\label{alg:ff} The Ford-Fulkerson Algorithm}

\textbf{Input:} $\mathcal{G}=(\mathcal{V},\mathcal{E}), \mathbf{C}, v_s, v_t$.

$f_{uv}=0: \forall (u,v)\in\mathcal{E}$

\textbf{while} $\exists\ \text{augmenting path}\ \mathcal{P}=\{v_s, \ldots, v_t\}\ \text{in}\ \mathcal{G}\ \text{do}:$

\quad $c_p=\min\{c_{uv}:(u,v)\in \mathcal{P}\}$ ;
 
\quad \textbf{for} each $(u,v)\in \mathcal{P}$ \textbf{do}:
 
\quad \quad $f_{uv}=f_{uv}-c_p$;
 
\quad \quad $f_{vu}=f_{vu}+c_p$;
 
\quad \quad $c_{uv}=c_{uv}-c_p$;
 
\quad \quad $c_{vu}=c_{vu}+c_p$;

\quad \textbf{end for}

\textbf{end while}

\textbf{Output:} $\mathbf{F}$.
\end{algorithm}

\subsubsection{The Neural Algorithmic Reasoning Model}
Before introducing the AIGNN architecture for the reconstructor and predictor, we first discuss how to infuse the algorithmic knowledge into a NAR model.
We use the Ford-Fulkerson algorithm to solve the max-flow problems, generating input-output pairs along with the full trajectory of the algorithm.
To avoid multiple valid shortest paths in the augmenting path-finding step, we generate a random weight matrix to select the shortest paths. 
For the neural execution of the algorithm, finding the shortest path with the Bellman-Ford algorithm can be achieved by learning to predict predecessors for each node \citep{velivckovic2020neural}.
The locations of the source and sink nodes are represented by an indicator vector, with two nonzero entries: 1 denotes the source node and -1 denotes the sink node.
There are five inputs to the NAR to characterize the max-flow problem: (1) the location indicator vector, (2) the capacity matrix, (3) the adjacency matrix, (4) a random weight matrix for finding the shortest path, and (5) a linearly spaced position feature from 0 to 1 for each node indexing the node.
The first four inputs are used to determine the problem instance, while the fifth input is introduced to serve as a position index of the node that can be used as a useful tie-breaker when algorithms could make an arbitrary choice on which node to explore next \citep{velivckovic2022clrs}.
The intermediate algorithmic trajectory is captured by hints, which include:
\begin{itemize}
    \item A mask vector indicating  whether the node is in the augmenting path or not
     \item The augmenting path represented by a vector specifying the predecessor node of each node
      \item The maximum flow $c_p$ that can pass through the current augmenting path $\mathcal{P}$
       \item The flow matrix $\mathbf{F}$.
\end{itemize}
The final output of the model is the flow matrix $\mathbf{F}$. 

As introduced in Section \ref{sec:NAR}, the NAR model consists of three parts: encoder, processor, and decoder. 
To ensure that the algorithmic knowledge is captured primarily by the processor, the encoders and decoders are chosen as linear layers. These layers have limited expressivity, contributing minimally to the learning of the algorithmic trajectory.
At each algorithm step, each input and the current hints are encoded with separate linear encoders. The encoded information is then processed by the processor, followed by decoders that generate the next hints or the algorithm output if it is the last algorithm step. 
It is important to note that there are two types of encoders in the NAR model. The first type encodes the inputs and is  used only once before the neural execution of the algorithm. The second type encodes the hints and is used in each algorithm step before feeding the node embeddings to the processor. 

In summary, the linear encoders and decoders ensure that the learning process focuses on the processor, which is designed to capture and replicate the algorithmic trajectory. This setup allows the NAR model to effectively learn and execute the desired algorithmic reasoning tasks.

As the core component of the NAR model, the choice of processor significantly influences its learning ability.
According to evaluation results in two benchmark studies \citep{velivckovic2022clrs,minder2023salsa}, the pointer graph networks (PGN) \citep{velivckovic2020pointer} outperform  various other models, including memory networks \citep{sukhbaatar2015end}, deep sets \citep{zaheer2017deep}, message-passing neural networks \citep{gilmer2017neural}, graph attention networks \citep{velivckovic2018graph}, graph isomorphism network \citep{xu2019powerful}, and recurrent GNNs \citep{grotschla2022learning}.
Therefore, we employ the PGN model as the processor in our NAR framework.
The PGN model consists of the following components: the source node layer $\boldsymbol{\Theta}_s$, the target node layer $\boldsymbol{\Theta}_t$, the edge layer $\boldsymbol{\Theta}_e$, the message transformation function $\boldsymbol{\Theta}_{msg}$, the skip connection layer $\boldsymbol{\Theta}_{skip}$ and the output layer $\boldsymbol{\Theta}_{out}$. 
Specifically, the message transformation function is defined as a two-layer MLP, while the other five layers are all defined as linear layers.
The update of node $v$ at the $k$-th step is defined as follows:
\begin{equation}
\begin{aligned}
\mathbf{h}_v^{k+1}&=\sigma\Biggl(\boldsymbol{\Theta}_{skip}(\mathbf{h}_v^k)  \\
 &  +\boldsymbol{\Theta}_{out}\Biggl(\underset{u\in\mathcal{N}(v)}{\bigoplus} \boldsymbol{\Theta}_{msg}\left(\boldsymbol{\Theta}_s(\mathbf{h}_v^k)+\boldsymbol{\Theta}_t(\mathbf{h}_u^k)+\boldsymbol{\Theta}_e(\mathbf{h}_{e_{uv}}^k)\right)\Biggr)\Biggr),
\end{aligned}
\end{equation}
where $\sigma$ represents the ReLU activation function and $\bigoplus$ denotes the elementwise max aggregation function. 
For algorithmic reasoning, max aggregation is more suitable than the commonly used sum aggregation because it aligns better with the computational mechanisms in classical algorithms.
It has been shown both empirically \citep{battaglia2018relational,velivckovic2020neural} and theoretically \citep{xu2020neural} that GNNs with max aggregation extrapolate better on certain algorithmic tasks (e.g., the shortest path finding problem) than GNNs with sum aggregation.
Additionally, when performing message passing, we treat the graph as an unweighted graph, and the random weight matrix for finding the shortest path is treated as edge features.

There are two subroutines in the Ford-Fulkerson algorithm: finding the shortest path and updating the capacity matrix $\mathbf{C}$ and the flow matrix $\mathbf{F}$.
Consequently, we use two subnetworks to serve as the processor.
The PGN model aligns with both subroutines and has the capability to learn and execute them effectively.
The possibility of learning multiple algorithms with a single processor has been explored in previous studies \citep{xhonneux2021transfer,ibarz2022generalist,Montgomery2024markov}. 
Motivated by their findings, we propose using a single PGN to perform both subroutines of the Ford-Fulkerson algorithm.
To inform the processor of which subroutine to execute, we include an additional binary feature.
Specifically, at each step, the PGN model is applied twice: first to compute the augmenting path $\mathcal{P}$, and then to update the capacity matrix $\mathbf{C}$ and the flow matrix $\mathbf{F}$.

To ensure the model behaves like the algorithm, we apply step-wise hint supervision.
The loss function comprises two components: the output loss and the step-wise hint loss.
After training, the processor is expected to emulate the Ford-Fulkerson algorithm. 
This trained processor can be extracted and applied to other scenarios where the algorithmic knowledge for computing max-flow is useful.

\subsubsection{Algorithm-Informed Reconstructor and Predictor}
After developing the NAR model capable of executing the Ford-Fulkerson algorithm, we transfer the learned algorithmic knowledge to GNNs for pressure reconstruction and predictions in WDNs,  creating the Algorithm-Informed Graph Neural Network (AIGNN).
Specifically, AIGNN replaces the encoder and decoder in the NAR with Chebyshev convolutional layers to handle the real-world inputs while retaining the processor.
The Chebyshev convolutional layer is defined by
\[
\mathbf{X}=\sum_{s=1}^S \mathbf{Z}^{(s)}\boldsymbol{\Theta}^{(s)},
\]
where $\mathbf{Z^{(s)}}$ is computed recursively by
\[
\begin{aligned}\mathbf{Z}^{(1)} & =\mathbf{X},\\
\mathbf{Z}^{(2)} & =\hat{\mathbf{L}}\mathbf{X},\\
\mathbf{Z}^{(s)} & =2\hat{\mathbf{L}}\mathbf{Z}^{(s-1)}-\mathbf{Z}^{(s-2)}
\end{aligned}
\]
with $\hat{\mathbf{Z}}=\frac{2\mathbf{L}}{\lambda_{\max}}-\mathbf{I}$ denoting the scaled and normalized Laplacian matrix.

Intuitively, the new encoder and decoder are used to map the real-world problem to and from the abstract space in which the processor operates in.
This requires the model to be more expressive compared to that used to map the max-flow problem to and from the abstract space.
Given that the processor's parameters are frozen, linear encoders and decoders lack the expressivity needed for effective mapping between the real world and the abstract space.
Thus, we choose Chebyshev convolution layers instead of linear layers for the encoders and decoders. 
Both the reconstructor and the predictor have the same architecture, differing only in their input dimensions.
The reconstructor AIGNN and the predictor AIGNN are trained independently, but they share the same processor adopted from the NAR model.

To train the newly introduced encoder and decoder, we use the partially observable pressure data from the WDN.
At timestep $t$, for the reconstructor, the sparse node pressure $\mathbf{x}(t)$ at time $t$ is used as the input, and we denote the outputs as $\mathbf{y}_r(t)$, representing the reconstructed pressure data for all nodes.
For the predictor, the sparse node pressures from previous $T$ timesteps $\mathbf{x}(t-T), \ldots, \mathbf{x}(t-1)$  are used as inputs, and we denote the outputs as $\mathbf{y}_p(t)$, representing the predicted pressure data for all nodes.
After training, $\hat{\mathbf{y}}_r(t)$ denotes the output of the reconstructor,  and  $\hat{\mathbf{y}}_p(t)$ denotes the output of the predictor at timestep $t$.
The parameters in the processor can either be frozen or fine-tuned along with the encoder and decoder.

Although the max-flow property can help in inferring pressure,  the processor's design to learn max-flow may result in the loss of other information that could contribute to pressure reconstruction and prediction.   
Therefore, in addition to using AIGNN directly for pressure reconstruction and prediction, we propose two variants that integrate the embeddings generated by AIGNN with other models.
This allows the resulting models to benefit from max-flow information without losing other useful information.
Specifically, we use the embedding before the last layer of AIGNN as the augmenting embedding. 
This embedding is then fed into the ChebNet model \citep{gardharsson2022graph} in two ways: as additional inputs (ChebNet$_\text{IN}$) or as additional embeddings before the last layer (ChebNet$_\text{EMB}$).

\subsection{Second Stage: Residual Analysis}
After obtaining the prediction and reconstruction pressures, we compute the residual pressure to identify the leakages.
At each time step $t$, a residual signal is computed based on the outputs of the reconstructor and the predictor, i.e., $\mathbf{r}(t)=\left[r_1(t),\ldots,r_N(t)\right]=\hat{\mathbf{y}}_{p}(t)-\hat{\mathbf{y}}_{r}(t)$, where the reconstructor is trained with data from time step $t$ and the predictor is trained with data from time step $t-T,\ldots,t-1$.
Since leakages occur at edges, while the residual signals are defined on nodes, we further compute the absolute edge residual as $r_{uv}^{(E)}(t)=\lvert r_v(t)-r_u(t)\rvert$.
Under normal conditions, the residual signals should mainly be caused by the random noise and have stationary behavior.
Thus, when the residual signals deviate from the stationary behavior, it implies that the WDN has some abnormal behavior.
To evaluate the deviation of the residual signals from its stationary point, we compute the moving average of $r_{uv}^{(E)}(t)$, denoted as $\bar{r}_{uv}^{(E)}(t)$.
At timestep $t$, with a window size of $s$, we have
\[
\bar{r}_{uv}^{(E)}(t)=\frac{1}{s}\sum_{i=0}^{s-1}r_{uv}^{(E)}(t-i).
\]
We further compute the mean of $\bar{r}_{uv}^{(E)}(t)$ over time period $1,\ldots,t_{\mathsf{max}}$ as
\[
\bar{r}_{uv}^{(E)} = \frac{1}{t_\mathsf{max}}\sum_{t=1}^{t_\mathsf{max}}\bar{r}_{uv}^{(E)}(t),
\]
and the corresponding standard deviation:
\[
\sigma_{uv}^{(E)} = \sqrt{\frac{1}{t_\mathsf{max}}\sum_{t=1}^{t_\mathsf{max}} \left(\bar{r}_{uv}^{(E)}(t) - \bar{r}_{uv}^{(E)}\right)^2}.
\]
When the moving average $\bar{r}_{uv}^{(E)}(t)$ exceeds a threshold $\bar{r}_{uv}^{(E)}+\xi \sigma_{uv}^{(E)}$ with parameter $\xi$, we mark it as a potential leakage.

\section{Case Study}\label{sec:experiments}
\subsection{Problem Setting}
\begin{figure*}
    \centering
    \includegraphics[width=1.0\textwidth]{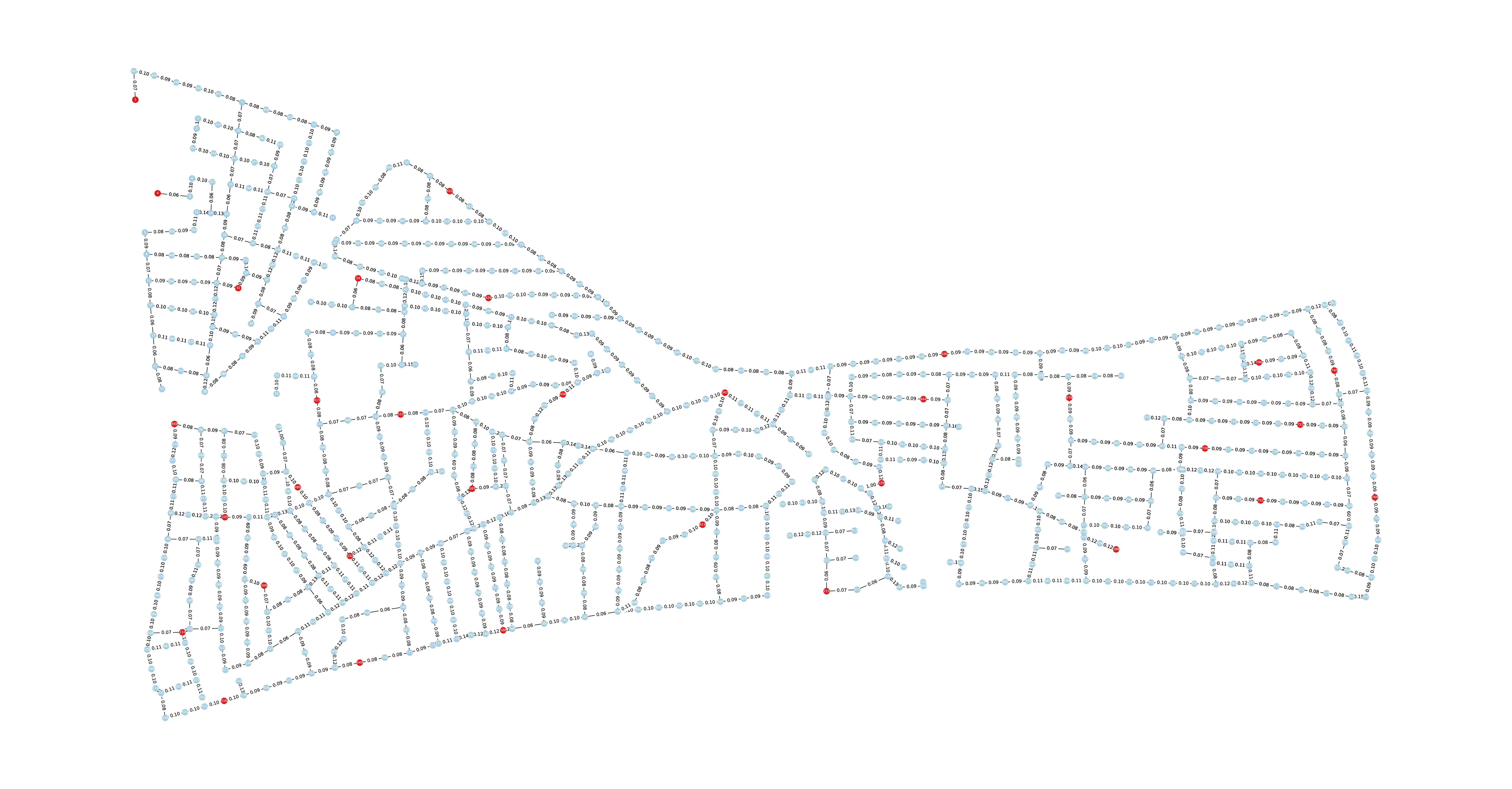}
    \vspace{-0.6cm}
    \caption{L-town network.\label{fig:l-town-network}}
\end{figure*}

In this section, we evaluate the performance of the proposed algorithm using a case study on the $L$-town WDN introduced in the BattleDIM competition \citep{vrachimis2020battledim}. 
The $L$-town WDN consists of 785 junctions and 905 pipes, with 33 pressure sensors located at different junctions across the network, collecting data with a 5-minute temporal resolution.
The WDN is modeled as a graph, where junctions are represented as nodes and pipes as edges.
The topology of the $L$-town WDN is illustrated in Figure \ref{fig:l-town-network}, where red nodes denote junctions with sensors and blue nodes denote junctions without sensors.
The $L$-town WDN also contains a nominal model represented with an EPANET hydraulic simulation input file, which has $\pm 10\%$ uncertainties to simulate a real-world noisy environment.
This nominal model is used to generate a training dataset for the reconstructor and predictor.
Additionally, the BattleDIM competition provides two years of data from 2018 and 2019.
Data from 2018 will be used to determine the threshold $\xi$ for residual analysis, while data from 2019 will be employed to evaluate the model's performance in leakage detection and localization.

We compare our method with the state-of-the-art GNN model for leakage detection and localization, specifically the ChebNet model proposed by  \citet{gardharsson2022graph}. 
Both approaches use a two-stage approach that first trains a reconstructor and predictor and then detects and localizes leakages by analyzing the residuals of these two networks' outputs.
The key difference between our approach and theirs lies in the architecture of the reconstructor and predictor. 
 The ChebNet model  uses the  architecture from \citet{hajgato2021reconstructing}, which consists of four Chebyshev spectral graph convolutional layers. The degree of the polynomial in the hidden layers is set to $[S_1,S_2,S_3]=[240,120,20]$, and the filter sizes are $[F_1,F_2,F_3]=[120,60,30]$.

\subsection{Training the AIGNN model}
To train the AIGNN model, we first pre-train a NAR model to execute the Ford-Fulkerson algorithm for solving max-flow problems.
This involves generating the algorithm trajectories of the Ford-Fulkerson algorithm for max-flow problems.
Each problem instance contains a graph sampled from the Erdos-Renyi distribution, with randomly assigned capacities on each edge.
The intermediate steps of the Ford-Fulkerson algorithm are represented with hints, and the output of the algorithm is represented by a flow matrix.
Specifically, following the setting in CLRS benchmark \citep{velivckovic2022clrs}, we sample 1000 graphs with 16 nodes each for training.
The NAR model is trained to predict the flow matrix and accurately recover all the intermediate steps of the algorithm.

After training the NAR model with Ford-Fulkerson algorithm instances, its processor inherits the algorithmic reasoning abilities.
We then adapt the processor to address the task of pressure reconstruction and prediction in WDNs.
Specifically, we replace the linear encoder and decoder of the NAR model with Chebyshev convolutional layers, finalizing the AIGNN model architecture.
Two AIGNN models are trained independently to address the pressure reconstruction and prediction tasks using training data obtained from the nominal model provided in \citep{vrachimis2020battledim}.

We consider three versions of AIGNN for pressure reconstruction and prediction:
\begin{itemize}
    \item AIGNN: The parameters of the processor inherited from the NAR model are fixed, while only the parameters of the encoder and decoder are trained with the pressure data.
    \item $\text{AIGNN}_{\text{FT}}$: The parameters of the processor are initialized with those from the NAR model but are fine-tuned together with the parameters in the encoder and decoder.
    \item  $\text{AIGNN}_{\text{POS}}$: This version has the same architecture and training strategy as AIGNN but includes additional positional features, computed as evenly spaced numbers between 0 and 1 according to the node index.
\end{itemize}

We compare the three models on the pressure reconstruction task, with the results presented in Table \ref{tab:NAR-pretrain}. 
The relative error is defined as 
\[\text{Rel. error} = \frac{\|\mathbf{y}_{r}(t)-\hat{\mathbf{y}}_{r}(t)\|}{\mathbf{y}_{r}(t)}.\]
where Rel. error (+) and Rel. error (-) represent the relative error at the monitored and unmonitored nodes, respectively.
From the table, we can  observe the following:
\begin{itemize}
    \item AIGNN provides the best overall relative error. This indicates that the fixed processor parameters, which incorporate the algorithmic knowledge from the NAR model, are sufficiently informative for the task. Further fine-tuning of the processor with real-world data does not lead to performance improvement.
    \item $\text{AIGNN}_{\text{FT}}$ where the processor is fine-tuned, does not perform as well as AIGNN. This suggests that the pre-trained algorithmic knowledge is more beneficial when left unchanged.

    \item $\text{AIGNN}_{\text{POS}}$, which includes additional positional features, does not show significant improvement for the pressure reconstruction tasks in WDNs. While positional features are helpful for algorithmic tasks as shown in the CLRS benchmark \citep{velivckovic2022clrs}, they do not appear to provide the same benefit here.
\end{itemize}
Given these observations, we will use AIGNN in the following experiments, as it demonstrated the best performance in Table \ref{tab:NAR-pretrain}.

\begin{table}[ht]
\begin{centering}
\caption{Relative reconstruction error of the reconstructor.}
\label{tab:NAR-pretrain}
\par\end{centering}
\centering{}\resizebox{1.0\columnwidth}{!}{%
\begin{tabular}{cccc}
\toprule 
\multirow{1}{*}{\%} & Rel. error & Rel. error (+) & Rel. error (-)\tabularnewline
\midrule 
AIGNN & 0.4092 $\pm$ 0.1032 & 0.4347 $\pm$ 0.0935 & 0.4081 $\pm$ 0.1042\tabularnewline
$\text{AIGNN}_{\text{FT}}$ & 0.4189 $\pm$ 0.1693 & 0.4279 $\pm$ 0.1596 & 0.4185 $\pm$ 0.1707\tabularnewline
$\text{AIGNN}_{\text{POS}}$ & 0.6526 $\pm$ 0.1335 & 0.6225 $\pm$ 0.1257 & 0.6539 $\pm$ 0.1341\tabularnewline
\bottomrule
\end{tabular}}
\end{table}

\subsection{Pressure Reconstruction and Prediction}
As demonstrated in Section \ref{sec:proposed_method}, our approach consists of two steps: (1) pressure reconstruction and prediction, and (2) residual analysis.
In this section, we first evaluate the performance of AIGNN on the pressure reconstruction and prediction tasks.
Since the processor in AIGNN is pre-trained with algorithmic trajectories and kept fixed afterward,  AIGNN mainly relies on the max-flow property of the network to infer the pressure.
Thus, we hypothesize that AIGNN extracts different semantic information from the pressure measurements compared to ChebNet.
Motivated by this, we propose two approaches that incorporate the information extracted by AIGNN into  ChebNet, namely ChebNet$_\text{IN}$ and ChebNet$_\text{EMB}$.
Specifically, ChebNet$_\text{IN}$ augments the AIGNN embedding as additional input while ChebNet$_\text{EMB}$ augments the AIGNN embedding as an additional embedding before the last layer.

The comparative results of the above models for pressure reconstruction and prediction are showcased in Table \ref{tab:reconstructor-result} and Table \ref{tab:predictor-result}, respectively.
From the results, we observe that AIGNN outperforms ChebNet in both pressure reconstruction and pressure prediction, indicating that the algorithmic knowledge embedded in AIGNN is more effective for inferring the pressure of the WDN.
The fact that both $\text{ChebNet}_{\text{IN}}$ and $\text{ChebNet}_{\text{EMB}}$ outperform AIGNN and ChebNet validates our hypothesis that the semantic information extracted by AIGNN differs from that extracted by ChebNet.
Moreover, the additional improvement achieved by $\text{ChebNet}_{\text{IN}}$ and $\text{ChebNet}_{\text{EMB}}$ indicates that AIGNN can effectively serve as an auxiliary model, enhancing the performance of existing models.

\begin{table}[ht]
\begin{centering}
\caption{Relative reconstruction error of the reconstructor.}
\label{tab:reconstructor-result}
\par\end{centering}
\centering{}\resizebox{1.0\columnwidth}{!}{%
\begin{tabular}{cccc}
\toprule 
\multirow{1}{*}{\%} & Rel. error & Rel. error (+) & Rel. error (-)\tabularnewline
\midrule 
ChebNet & 0.4575\textbf{ $\pm$ }0.1067 & 0.4280\textbf{ $\pm$ }0.1144 & 0.4588\textbf{ $\pm$ }0.1069\tabularnewline
AIGNN & 0.4092 $\pm$ 0.1032 & 0.4347 $\pm$ 0.0935 & 0.4081 $\pm$ 0.1042\tabularnewline
$\text{ChebNet}_{\text{IN}}$ & 0.3645 $\pm$ 0.1307 & 0.3433 $\pm$ 0.1348 & 0.3654 $\pm$ 0.1306\tabularnewline
$\text{ChebNet}_{\text{EMB}}$ & 0.2392 $\pm$ 0.0339 & 0.2342 $\pm$ 0.0375 & 0.2394 $\pm$ 0.0341\tabularnewline
\bottomrule
\end{tabular}}
\end{table}

\begin{table}[ht]
\begin{centering}
\caption{Relative prediction error of the predictor.}
\label{tab:predictor-result}
\par\end{centering}
\centering{}\resizebox{1.0\columnwidth}{!}{%
\begin{tabular}{cccc}
\toprule 
\multirow{1}{*}{\%} & Rel. error & Rel. error (+) & Rel. error (-)\tabularnewline
\midrule 
ChebNet & 0.3774\textbf{ $\pm$ }0.1588 & 0.3905\textbf{ $\pm$ }0.1784 & 0.3769\textbf{ $\pm$ }0.1581\tabularnewline
AIGNN & 0.3249\textbf{ $\pm$ }0.0685 & 0.3271\textbf{ $\pm$ }0.0633 & 0.3248\textbf{ $\pm$ }0.0691\tabularnewline
$\text{ChebNet}_{\text{IN}}$ & 0.2964\textbf{ $\pm$ }0.0874 & 0.2877\textbf{ $\pm$ }0.0886 & 0.2968\textbf{ $\pm$ }0.0874\tabularnewline
$\text{ChebNet}_{\text{EMB}}$ & 0.2077\textbf{ $\pm$ }0.0074 & 0.2126\textbf{ $\pm$ }0.0153 & 0.2075\textbf{ $\pm$ }0.0073\tabularnewline
\bottomrule
\end{tabular}}
\end{table}

\subsection{Evaluation of the generalization ability}
To validate the generalization ability of the models, we conduct experiments on different configurations of sensor locations, training the models on only one sensor configuration. Standard approaches typically fail to generalize in such setups. We hypothesize that the processor in our approach may have captured more general information about the system flow, enabling it to transfer effectively to other sensor configurations. 
Specifically, we first train the model using the default sensor placement provided in the dataset.
Then, we test the model on scenarios with random sensor placement, maintaining the same number of sensors.
We conduct experiments with five different sensor placements and report the average results with standard deviation in Table \ref{tab:sensor_placement}
The results indicate that the three algorithm-informed models, i.e., AIGNN, ChebNet$_\text{IN}$, and ChebNet$_\text{EMB}$, perform better than the ChebNet model, suggesting that algorithmic knowledge positively contributes to generalization capabilities of the models.
Among these three algorithm-informed models, ChebNet$_\text{EMB}$ performs the worst, as it only incorporates the algorithm information before the last layer, leaving most of the reasoning process to the non-algorithmic part.
Conversely, ChebNet$_\text{IN}$ achieves the best generalization performance since it uses the AIGNN embedding as input, forcing the model to focus more on the generalizable algorithmic information.
The promising results obtained with ChebNet$_\text{IN}$ also imply the potential of using AIGNN as an auxiliary enhancement for other existing methods.

\begin{table}[ht]
\begin{centering}
\caption{Relative reconstruction error with different sensor configurations.}
\label{tab:sensor_placement}
\par\end{centering}
\centering{}\resizebox{1.0\columnwidth}{!}{%
\begin{tabular}{cccc}
\toprule 
\multirow{1}{*}{\%} & Rel. error & Rel. error (+) & Rel. error (-)\tabularnewline
\midrule 
ChebNet & 17.065\textbf{ $\pm$ }1.7626 & 14.778\textbf{ $\pm$ }2.9329 & 17.165\textbf{ $\pm$ }1.7125\tabularnewline
AIGNN & 9.3255\textbf{ $\pm$ }0.6601 & 9.0211\textbf{ $\pm$ }2.2847 & 9.3389\textbf{ $\pm$ }0.6106\tabularnewline
$\text{ChebNet}_{\text{IN}}$ & 2.2823\textbf{ $\pm$ }0.5074 & 3.1819\textbf{ $\pm$ }0.7282 & 2.2427\textbf{ $\pm$ }0.5077\tabularnewline
$\text{ChebNet}_{\text{EMB}}$ & 11.085\textbf{ $\pm$ }1.3645 & 11.449\textbf{ $\pm$ }2.7778 & 11.069\textbf{ $\pm$ }1.3981\tabularnewline
\bottomrule
\end{tabular}}
\end{table}
\subsection{Leakage Detection}
To detect and localize the leakages, we first calculate the edge residual signal $\bar{r}_{uv}^{(E)}(t)$.
Using data from 2018, during which 14 leakages occurred, we determine the threshold parameter $\xi$.
The sliding window size $s$ is set to $12$, covering one hour.
To tune the threshold parameter $\xi$, we initialize it as 3 and decrease it with intervals of 0.05 until at least 12 out of 14 leakages are detected.
This parameter-tuning scheme is applied across all models to ensure a fair comparison, which allows us to evaluate different models’ ability to detect leakages.
To reduce false alarms, a leakage is only reported if the residuals for the corresponding pipe exceed the threshold consecutively for six hours.
After parameter-tuning using the data from 2018, we test the model using the data from 2019, during which 23 leakages occurred.
The number of detected leakages along with the corresponding values of $\xi$, are summarized in Table \ref{tab:leak_detection}.
The results indicate that $\text{ChebNet}_{\text{IN}}$ and $\text{ChebNet}_{\text{EMB}}$ detect more leakages compared to ChebNet, demonstrating the benefits of incorporating additional algorithmic knowledge.
These two models also exhibit larger values of $\xi$, suggesting that leakages are more discernible in $\text{ChebNet}_{\text{IN}}$ and $\text{ChebNet}_{\text{EMB}}$ compared to ChebNet and AIGNN. 
Although AIGNN had the most detected leakages in 2019, it is mainly due to its low $\xi$ value, which also resulted in numerous false alarms.
The results highlight the importance of combining both algorithmic and non-algorithmic knowledge for effective leakage detection.
\begin{table}[ht]
\begin{centering}
\caption{Leakage detection performance comparison.\label{tab:leak_detection}}
\par\end{centering}
\centering{}\resizebox{1.0\columnwidth}{!}{%
\begin{tabular}{cccc}
\toprule 
 & $\xi$ & Deteced leakages (2018) & Detected leakages (2019) \tabularnewline
\midrule 
ChebNet & 1.2 & 12/14 & 15/23\tabularnewline
AIGNN & 0.25 & 12/14 & 22/23\tabularnewline
$\text{ChebNet}_{\text{IN}}$ & 1.4 & 12/14 & 17/23\tabularnewline
$\text{ChebNet}_{\text{EMB}}$ & 1.6 & 12/14 & 19/23\tabularnewline
\bottomrule
\end{tabular}}
\end{table}

\section{Conclusions}\label{sec:conclusions}
In this paper, we propose Algorithm-Informed Graph Neural Networks (AIGNN) to address the leakage detection and localization problem in WDNs by leveraging Ford-Fulkerson algorithmic knowledge for solving max-flow problems.
Beyond using AIGNN to directly tackle leakage detection and localization, its embeddings can also augment the input or intermediate embedding to existing GNN models.
A case study on the $L$-town network demonstrates the effectiveness and generalization ability of AIGNN.
Integrating ChebNet with AIGNN results in additional performance and generalization improvements, indicating the potential of using algorithmic-informed models as auxiliary components in existing frameworks.
This paper makes a significant step toward transferring algorithmic knowledge to solve engineering problems. 
The proposed approach can be easily transferred to help solve problems in other flow networks, such as transportation networks and communication networks.
Additionally, the results suggest promising prospects for algorithm-informed neural networks in engineering applications, facilitating the development of more generalizable models and opening up new potential areas for future explorations.

\section*{Acknowledgement}
This work has been supported by the Swiss National Science Foundation (SNSF) Grant 200021\_200461.
\bibliographystyle{elsarticle-harv}
\bibliography{ref}

\begin{thebibliography}{60}
\expandafter\ifx\csname natexlab\endcsname\relax\def\natexlab#1{#1}\fi
\providecommand{\url}[1]{\texttt{#1}}
\providecommand{\href}[2]{#2}
\providecommand{\path}[1]{#1}
\providecommand{\DOIprefix}{doi:}
\providecommand{\ArXivprefix}{arXiv:}
\providecommand{\URLprefix}{URL: }
\providecommand{\Pubmedprefix}{pmid:}
\providecommand{\doi}[1]{\href{http://dx.doi.org/#1}{\path{#1}}}
\providecommand{\Pubmed}[1]{\href{pmid:#1}{\path{#1}}}
\providecommand{\bibinfo}[2]{#2}
\ifx\xfnm\relax \def\xfnm[#1]{\unskip,\space#1}\fi
\bibitem[{Ashraf et~al.(2023)Ashraf, Hermes, Artelt and Hammer}]{ashraf2023spatial}
\bibinfo{author}{Ashraf, I.}, \bibinfo{author}{Hermes, L.}, \bibinfo{author}{Artelt, A.}, \bibinfo{author}{Hammer, B.}, \bibinfo{year}{2023}.
\newblock \bibinfo{title}{Spatial graph convolution neural networks for water distribution systems}, in: \bibinfo{booktitle}{International Symposium on Intelligent Data Analysis}, \bibinfo{organization}{Springer}. pp. \bibinfo{pages}{29--41}.
\bibitem[{Battaglia et~al.(2018)Battaglia, Hamrick, Bapst, Sanchez-Gonzalez, Zambaldi, Malinowski, Tacchetti, Raposo, Santoro, Faulkner et~al.}]{battaglia2018relational}
\bibinfo{author}{Battaglia, P.W.}, \bibinfo{author}{Hamrick, J.B.}, \bibinfo{author}{Bapst, V.}, \bibinfo{author}{Sanchez-Gonzalez, A.}, \bibinfo{author}{Zambaldi, V.}, \bibinfo{author}{Malinowski, M.}, \bibinfo{author}{Tacchetti, A.}, \bibinfo{author}{Raposo, D.}, \bibinfo{author}{Santoro, A.}, \bibinfo{author}{Faulkner, R.}, et~al., \bibinfo{year}{2018}.
\newblock \bibinfo{title}{Relational inductive biases, deep learning, and graph networks}.
\newblock \bibinfo{journal}{arXiv preprint arXiv:1806.01261} .
\bibitem[{Bellman(1958)}]{bellman1958routing}
\bibinfo{author}{Bellman, R.}, \bibinfo{year}{1958}.
\newblock \bibinfo{title}{On a routing problem}.
\newblock \bibinfo{journal}{Quarterly of applied mathematics} \bibinfo{volume}{16}, \bibinfo{pages}{87--90}.
\bibitem[{Beurer-Kellner et~al.(2022)Beurer-Kellner, Vechev, Vanbever and Veli{\v{c}}kovi{\'c}}]{beurer2022learning}
\bibinfo{author}{Beurer-Kellner, L.}, \bibinfo{author}{Vechev, M.}, \bibinfo{author}{Vanbever, L.}, \bibinfo{author}{Veli{\v{c}}kovi{\'c}, P.}, \bibinfo{year}{2022}.
\newblock \bibinfo{title}{Learning to configure computer networks with neural algorithmic reasoning}.
\newblock \bibinfo{journal}{Advances in Neural Information Processing Systems} \bibinfo{volume}{35}, \bibinfo{pages}{730--742}.
\bibitem[{Bevilacqua et~al.(2023)Bevilacqua, Nikiforou, Ibarz, Bica, Paganini, Blundell, Mitrovic and Veli{\v{c}}kovi{\'c}}]{bevilacqua2023neural}
\bibinfo{author}{Bevilacqua, B.}, \bibinfo{author}{Nikiforou, K.}, \bibinfo{author}{Ibarz, B.}, \bibinfo{author}{Bica, I.}, \bibinfo{author}{Paganini, M.}, \bibinfo{author}{Blundell, C.}, \bibinfo{author}{Mitrovic, J.}, \bibinfo{author}{Veli{\v{c}}kovi{\'c}, P.}, \bibinfo{year}{2023}.
\newblock \bibinfo{title}{Neural algorithmic reasoning with causal regularisation}, in: \bibinfo{booktitle}{International Conference on Machine Learning}, \bibinfo{organization}{PMLR}.
\bibitem[{Cappart et~al.(2023)Cappart, Ch{\'e}telat, Khalil, Lodi, Morris and Veli{\v{c}}kovi{\'c}}]{cappart2023combinatorial}
\bibinfo{author}{Cappart, Q.}, \bibinfo{author}{Ch{\'e}telat, D.}, \bibinfo{author}{Khalil, E.B.}, \bibinfo{author}{Lodi, A.}, \bibinfo{author}{Morris, C.}, \bibinfo{author}{Veli{\v{c}}kovi{\'c}, P.}, \bibinfo{year}{2023}.
\newblock \bibinfo{title}{Combinatorial optimization and reasoning with graph neural networks}.
\newblock \bibinfo{journal}{Journal of Machine Learning Research} \bibinfo{volume}{24}, \bibinfo{pages}{1--61}.
\bibitem[{Chan et~al.(2018)Chan, Chin and Zhong}]{chan2018review}
\bibinfo{author}{Chan, T.K.}, \bibinfo{author}{Chin, C.S.}, \bibinfo{author}{Zhong, X.}, \bibinfo{year}{2018}.
\newblock \bibinfo{title}{Review of current technologies and proposed intelligent methodologies for water distributed network leakage detection}.
\newblock \bibinfo{journal}{Ieee Access} \bibinfo{volume}{6}, \bibinfo{pages}{78846--78867}.
\bibitem[{Cormen et~al.(2022)Cormen, Leiserson, Rivest and Stein}]{cormen2022introduction}
\bibinfo{author}{Cormen, T.H.}, \bibinfo{author}{Leiserson, C.E.}, \bibinfo{author}{Rivest, R.L.}, \bibinfo{author}{Stein, C.}, \bibinfo{year}{2022}.
\newblock \bibinfo{title}{Introduction to algorithms}.
\newblock \bibinfo{publisher}{MIT press}.
\bibitem[{Cuguer{\'o}-Escofet et~al.(2016)Cuguer{\'o}-Escofet, Quevedo, Alippi, Roveri, Puig, Garc{\'\i}a and Trov{\`o}}]{cuguero2016model}
\bibinfo{author}{Cuguer{\'o}-Escofet, M.{\`A}.}, \bibinfo{author}{Quevedo, J.}, \bibinfo{author}{Alippi, C.}, \bibinfo{author}{Roveri, M.}, \bibinfo{author}{Puig, V.}, \bibinfo{author}{Garc{\'\i}a, D.}, \bibinfo{author}{Trov{\`o}, F.}, \bibinfo{year}{2016}.
\newblock \bibinfo{title}{Model-vs. data-based approaches applied to fault diagnosis in potable water supply networks}.
\newblock \bibinfo{journal}{Journal of Hydroinformatics} \bibinfo{volume}{18}, \bibinfo{pages}{831--850}.
\bibitem[{Dudzik and Veli{\v{c}}kovi{\'c}(2022)}]{dudzik2022graph}
\bibinfo{author}{Dudzik, A.J.}, \bibinfo{author}{Veli{\v{c}}kovi{\'c}, P.}, \bibinfo{year}{2022}.
\newblock \bibinfo{title}{Graph neural networks are dynamic programmers}.
\newblock \bibinfo{journal}{Advances in Neural Information Processing Systems} \bibinfo{volume}{35}, \bibinfo{pages}{20635--20647}.
\bibitem[{Fang et~al.(2019)Fang, Zhang, Xie and Yang}]{fang2019detection}
\bibinfo{author}{Fang, Q.}, \bibinfo{author}{Zhang, J.}, \bibinfo{author}{Xie, C.}, \bibinfo{author}{Yang, Y.}, \bibinfo{year}{2019}.
\newblock \bibinfo{title}{Detection of multiple leakage points in water distribution networks based on convolutional neural networks}.
\newblock \bibinfo{journal}{Water Supply} \bibinfo{volume}{19}, \bibinfo{pages}{2231--2239}.
\bibitem[{Ford(1956)}]{ford1956network}
\bibinfo{author}{Ford, L.}, \bibinfo{year}{1956}.
\newblock \bibinfo{title}{Network flow theory}.
\newblock \bibinfo{journal}{Rand Corporation Paper, Santa Monica, 1956} .
\bibitem[{Ford and Fulkerson(1956)}]{ford1956maximal}
\bibinfo{author}{Ford, L.R.}, \bibinfo{author}{Fulkerson, D.R.}, \bibinfo{year}{1956}.
\newblock \bibinfo{title}{Maximal flow through a network}.
\newblock \bibinfo{journal}{Canadian journal of Mathematics} \bibinfo{volume}{8}, \bibinfo{pages}{399--404}.
\bibitem[{Gar{\dh}arsson et~al.(2022)Gar{\dh}arsson, Boem and Toni}]{gardharsson2022graph}
\bibinfo{author}{Gar{\dh}arsson, G.{\"O}.}, \bibinfo{author}{Boem, F.}, \bibinfo{author}{Toni, L.}, \bibinfo{year}{2022}.
\newblock \bibinfo{title}{Graph-based learning for leak detection and localisation in water distribution networks}.
\newblock \bibinfo{journal}{IFAC-PapersOnLine} \bibinfo{volume}{55}, \bibinfo{pages}{661--666}.
\bibitem[{Georgiev and Li{\'o}(2020)}]{georgiev2020neural}
\bibinfo{author}{Georgiev, D.}, \bibinfo{author}{Li{\'o}, P.}, \bibinfo{year}{2020}.
\newblock \bibinfo{title}{Neural bipartite matching}.
\newblock \bibinfo{journal}{arXiv preprint arXiv:2005.11304} .
\bibitem[{Georgiev et~al.(2023)Georgiev, Vinas, Considine, Dumitrascu and Lio}]{georgiev2023narti}
\bibinfo{author}{Georgiev, D.}, \bibinfo{author}{Vinas, R.}, \bibinfo{author}{Considine, S.}, \bibinfo{author}{Dumitrascu, B.}, \bibinfo{author}{Lio, P.}, \bibinfo{year}{2023}.
\newblock \bibinfo{title}{Narti: Neural algorithmic reasoning for trajectory inference}, in: \bibinfo{booktitle}{The 2023 ICML Workshop on Computational Biology}.
\bibitem[{Gilmer et~al.(2017)Gilmer, Schoenholz, Riley, Vinyals and Dahl}]{gilmer2017neural}
\bibinfo{author}{Gilmer, J.}, \bibinfo{author}{Schoenholz, S.S.}, \bibinfo{author}{Riley, P.F.}, \bibinfo{author}{Vinyals, O.}, \bibinfo{author}{Dahl, G.E.}, \bibinfo{year}{2017}.
\newblock \bibinfo{title}{Neural message passing for quantum chemistry}, in: \bibinfo{booktitle}{International conference on machine learning}, \bibinfo{organization}{PMLR}. pp. \bibinfo{pages}{1263--1272}.
\bibitem[{Gr{\"o}tschla et~al.(2022)Gr{\"o}tschla, Mathys and Wattenhofer}]{grotschla2022learning}
\bibinfo{author}{Gr{\"o}tschla, F.}, \bibinfo{author}{Mathys, J.}, \bibinfo{author}{Wattenhofer, R.}, \bibinfo{year}{2022}.
\newblock \bibinfo{title}{Learning graph algorithms with recurrent graph neural networks}.
\newblock \bibinfo{journal}{arXiv preprint arXiv:2212.04934} .
\bibitem[{Gupta et~al.(2017)Gupta, Bokde, Marathe and Kulat}]{gupta2017leakage}
\bibinfo{author}{Gupta, A.}, \bibinfo{author}{Bokde, N.}, \bibinfo{author}{Marathe, D.}, \bibinfo{author}{Kulat, K.}, \bibinfo{year}{2017}.
\newblock \bibinfo{title}{Leakage reduction in water distribution systems with efficient placement and control of pressure reducing valves using soft computing techniques}.
\newblock \bibinfo{journal}{Engineering, Technology \& Applied Science Research} \bibinfo{volume}{7}, \bibinfo{pages}{1528--1534}.
\bibitem[{Hajgat{\'o} et~al.(2021)Hajgat{\'o}, Gyires-T{\'o}th and Pa{\'a}l}]{hajgato2021reconstructing}
\bibinfo{author}{Hajgat{\'o}, G.}, \bibinfo{author}{Gyires-T{\'o}th, B.}, \bibinfo{author}{Pa{\'a}l, G.}, \bibinfo{year}{2021}.
\newblock \bibinfo{title}{Reconstructing nodal pressures in water distribution systems with graph neural networks}.
\newblock \bibinfo{journal}{arXiv preprint arXiv:2104.13619} .
\bibitem[{Hamrick et~al.(2018)Hamrick, Allen, Bapst, Zhu, McKee, Tenenbaum and Battaglia}]{hamrick2018relational}
\bibinfo{author}{Hamrick, J.B.}, \bibinfo{author}{Allen, K.R.}, \bibinfo{author}{Bapst, V.}, \bibinfo{author}{Zhu, T.}, \bibinfo{author}{McKee, K.R.}, \bibinfo{author}{Tenenbaum, J.B.}, \bibinfo{author}{Battaglia, P.W.}, \bibinfo{year}{2018}.
\newblock \bibinfo{title}{Relational inductive bias for physical construction in humans and machines}.
\newblock \bibinfo{journal}{arXiv preprint arXiv:1806.01203} .
\bibitem[{Harris and Ross(1955)}]{harris1955fundamentals}
\bibinfo{author}{Harris, T.}, \bibinfo{author}{Ross, F.}, \bibinfo{year}{1955}.
\newblock \bibinfo{title}{Fundamentals of a method for evaluating rail net capacities}.
\newblock \bibinfo{type}{Technical Report}. Rand Corporation.
\bibitem[{Ibarz et~al.(2022)Ibarz, Kurin, Papamakarios, Nikiforou, Bennani, Csord{\'a}s, Dudzik, Bo{\v{s}}njak, Vitvitskyi, Rubanova et~al.}]{ibarz2022generalist}
\bibinfo{author}{Ibarz, B.}, \bibinfo{author}{Kurin, V.}, \bibinfo{author}{Papamakarios, G.}, \bibinfo{author}{Nikiforou, K.}, \bibinfo{author}{Bennani, M.}, \bibinfo{author}{Csord{\'a}s, R.}, \bibinfo{author}{Dudzik, A.J.}, \bibinfo{author}{Bo{\v{s}}njak, M.}, \bibinfo{author}{Vitvitskyi, A.}, \bibinfo{author}{Rubanova, Y.}, et~al., \bibinfo{year}{2022}.
\newblock \bibinfo{title}{A generalist neural algorithmic learner}, in: \bibinfo{booktitle}{Learning on Graphs Conference}, \bibinfo{organization}{PMLR}. pp. \bibinfo{pages}{2--1}.
\bibitem[{Islam et~al.(2022)Islam, Azam, Shanmugam and Mathur}]{islam2022review}
\bibinfo{author}{Islam, M.R.}, \bibinfo{author}{Azam, S.}, \bibinfo{author}{Shanmugam, B.}, \bibinfo{author}{Mathur, D.}, \bibinfo{year}{2022}.
\newblock \bibinfo{title}{A review on current technologies and future direction of water leakage detection in water distribution network}.
\newblock \bibinfo{journal}{IEEE Access} .
\bibitem[{Laucelli et~al.(2016)Laucelli, Romano, Savi{\'c} and Giustolisi}]{laucelli2016detecting}
\bibinfo{author}{Laucelli, D.}, \bibinfo{author}{Romano, M.}, \bibinfo{author}{Savi{\'c}, D.}, \bibinfo{author}{Giustolisi, O.}, \bibinfo{year}{2016}.
\newblock \bibinfo{title}{Detecting anomalies in water distribution networks using epr modelling paradigm}.
\newblock \bibinfo{journal}{Journal of Hydroinformatics} \bibinfo{volume}{18}, \bibinfo{pages}{409--427}.
\bibitem[{Li et~al.(2024)Li, Liu, Zhang and Fu}]{li2024real}
\bibinfo{author}{Li, Z.}, \bibinfo{author}{Liu, H.}, \bibinfo{author}{Zhang, C.}, \bibinfo{author}{Fu, G.}, \bibinfo{year}{2024}.
\newblock \bibinfo{title}{Real-time water quality prediction in water distribution networks using graph neural networks with sparse monitoring data}.
\newblock \bibinfo{journal}{Water Research} \bibinfo{volume}{250}, \bibinfo{pages}{121018}.
\bibitem[{Liemberger and Wyatt(2019)}]{liemberger2019quantifying}
\bibinfo{author}{Liemberger, R.}, \bibinfo{author}{Wyatt, A.}, \bibinfo{year}{2019}.
\newblock \bibinfo{title}{Quantifying the global non-revenue water problem}.
\newblock \bibinfo{journal}{Water Supply} \bibinfo{volume}{19}, \bibinfo{pages}{831--837}.
\bibitem[{de~Luca and Fountoulakis(2024)}]{de2024simulation}
\bibinfo{author}{de~Luca, A.B.}, \bibinfo{author}{Fountoulakis, K.}, \bibinfo{year}{2024}.
\newblock \bibinfo{title}{Simulation of graph algorithms with looped transformers}.
\newblock \bibinfo{journal}{arXiv preprint arXiv:2402.01107} .
\bibitem[{Mahdavi et~al.(2023)Mahdavi, Swersky, Kipf, Hashemi, Thrampoulidis and Liao}]{mahdavi2023towards}
\bibinfo{author}{Mahdavi, S.}, \bibinfo{author}{Swersky, K.}, \bibinfo{author}{Kipf, T.}, \bibinfo{author}{Hashemi, M.}, \bibinfo{author}{Thrampoulidis, C.}, \bibinfo{author}{Liao, R.}, \bibinfo{year}{2023}.
\newblock \bibinfo{title}{Towards better out-of-distribution generalization of neural algorithmic reasoning tasks}.
\newblock \bibinfo{journal}{Transactions on Machine Learning Research} .
\bibitem[{Minder et~al.(2023)Minder, Gr{\"o}tschla, Mathys and Wattenhofer}]{minder2023salsa}
\bibinfo{author}{Minder, J.}, \bibinfo{author}{Gr{\"o}tschla, F.}, \bibinfo{author}{Mathys, J.}, \bibinfo{author}{Wattenhofer, R.}, \bibinfo{year}{2023}.
\newblock \bibinfo{title}{Salsa-clrs: A sparse and scalable benchmark for algorithmic reasoning}, in: \bibinfo{booktitle}{The Second Learning on Graphs Conference}.
\bibitem[{Montgomery et~al.(2024)Montgomery, Meng, Alexandra and Shuiwang}]{Montgomery2024markov}
\bibinfo{author}{Montgomery, B.}, \bibinfo{author}{Meng, L.}, \bibinfo{author}{Alexandra, S.}, \bibinfo{author}{Shuiwang, J.}, \bibinfo{year}{2024}.
\newblock \bibinfo{title}{On the markov property of neural algorithmic reasoning: Analyses and methods}, in: \bibinfo{booktitle}{The Twelfth International Conference on Learning Representations}.
\bibitem[{Mounce et~al.(2011)Mounce, Mounce and Boxall}]{mounce2011novelty}
\bibinfo{author}{Mounce, S.R.}, \bibinfo{author}{Mounce, R.B.}, \bibinfo{author}{Boxall, J.B.}, \bibinfo{year}{2011}.
\newblock \bibinfo{title}{Novelty detection for time series data analysis in water distribution systems using support vector machines}.
\newblock \bibinfo{journal}{Journal of hydroinformatics} \bibinfo{volume}{13}, \bibinfo{pages}{672--686}.
\bibitem[{Numeroso et~al.(2023)Numeroso, Bacciu and Veli{\v{c}}kovi{\'c}}]{numeroso2023dual}
\bibinfo{author}{Numeroso, D.}, \bibinfo{author}{Bacciu, D.}, \bibinfo{author}{Veli{\v{c}}kovi{\'c}, P.}, \bibinfo{year}{2023}.
\newblock \bibinfo{title}{Dual algorithmic reasoning}, in: \bibinfo{booktitle}{The Eleventh International Conference on Learning Representations}.
\bibitem[{Puust et~al.(2010)Puust, Kapelan, Savic and Koppel}]{puust2010review}
\bibinfo{author}{Puust, R.}, \bibinfo{author}{Kapelan, Z.}, \bibinfo{author}{Savic, D.}, \bibinfo{author}{Koppel, T.}, \bibinfo{year}{2010}.
\newblock \bibinfo{title}{A review of methods for leakage management in pipe networks}.
\newblock \bibinfo{journal}{Urban Water Journal} \bibinfo{volume}{7}, \bibinfo{pages}{25--45}.
\bibitem[{Rodionov and Prokhorenkova(2024)}]{rodionov2024discrete}
\bibinfo{author}{Rodionov, G.}, \bibinfo{author}{Prokhorenkova, L.}, \bibinfo{year}{2024}.
\newblock \bibinfo{title}{Discrete neural algorithmic reasoning}.
\newblock \bibinfo{journal}{arXiv preprint arXiv:2402.11628} .
\bibitem[{Romero-Ben et~al.(2022)Romero-Ben, Alves, Blesa, Cembrano, Puig and Duviella}]{romero2022leak}
\bibinfo{author}{Romero-Ben, L.}, \bibinfo{author}{Alves, D.}, \bibinfo{author}{Blesa, J.}, \bibinfo{author}{Cembrano, G.}, \bibinfo{author}{Puig, V.}, \bibinfo{author}{Duviella, E.}, \bibinfo{year}{2022}.
\newblock \bibinfo{title}{Leak localization in water distribution networks using data-driven and model-based approaches}.
\newblock \bibinfo{journal}{Journal of Water Resources Planning and Management} \bibinfo{volume}{148}, \bibinfo{pages}{04022016}.
\bibitem[{Romero-Ben et~al.(2023)Romero-Ben, Alves, Blesa, Cembrano, Puig and Duviella}]{romero2023leak}
\bibinfo{author}{Romero-Ben, L.}, \bibinfo{author}{Alves, D.}, \bibinfo{author}{Blesa, J.}, \bibinfo{author}{Cembrano, G.}, \bibinfo{author}{Puig, V.}, \bibinfo{author}{Duviella, E.}, \bibinfo{year}{2023}.
\newblock \bibinfo{title}{Leak detection and localization in water distribution networks: Review and perspective}.
\newblock \bibinfo{journal}{Annual Reviews in Control} .
\bibitem[{Sanz et~al.(2016)Sanz, P{\'e}rez, Kapelan and Savic}]{sanz2016leak}
\bibinfo{author}{Sanz, G.}, \bibinfo{author}{P{\'e}rez, R.}, \bibinfo{author}{Kapelan, Z.}, \bibinfo{author}{Savic, D.}, \bibinfo{year}{2016}.
\newblock \bibinfo{title}{Leak detection and localization through demand components calibration}.
\newblock \bibinfo{journal}{Journal of Water Resources Planning and Management} \bibinfo{volume}{142}, \bibinfo{pages}{04015057}.
\bibitem[{Soldevila et~al.(2020)Soldevila, Blesa, Jensen, Tornil-Sin, Fern{\'a}ndez-Cant{\'\i} and Puig}]{soldevila2020leak}
\bibinfo{author}{Soldevila, A.}, \bibinfo{author}{Blesa, J.}, \bibinfo{author}{Jensen, T.N.}, \bibinfo{author}{Tornil-Sin, S.}, \bibinfo{author}{Fern{\'a}ndez-Cant{\'\i}, R.M.}, \bibinfo{author}{Puig, V.}, \bibinfo{year}{2020}.
\newblock \bibinfo{title}{Leak localization method for water-distribution networks using a data-driven model and dempster--shafer reasoning}.
\newblock \bibinfo{journal}{IEEE transactions on control systems technology} \bibinfo{volume}{29}, \bibinfo{pages}{937--948}.
\bibitem[{Sophocleous et~al.(2019)Sophocleous, Savi{\'c} and Kapelan}]{sophocleous2019leak}
\bibinfo{author}{Sophocleous, S.}, \bibinfo{author}{Savi{\'c}, D.}, \bibinfo{author}{Kapelan, Z.}, \bibinfo{year}{2019}.
\newblock \bibinfo{title}{Leak localization in a real water distribution network based on search-space reduction}.
\newblock \bibinfo{journal}{Journal of Water Resources Planning and Management} \bibinfo{volume}{145}, \bibinfo{pages}{04019024}.
\bibitem[{Steffelbauer et~al.(2022)Steffelbauer, Deuerlein, Gilbert, Abraham and Piller}]{steffelbauer2022pressure}
\bibinfo{author}{Steffelbauer, D.B.}, \bibinfo{author}{Deuerlein, J.}, \bibinfo{author}{Gilbert, D.}, \bibinfo{author}{Abraham, E.}, \bibinfo{author}{Piller, O.}, \bibinfo{year}{2022}.
\newblock \bibinfo{title}{Pressure-leak duality for leak detection and localization in water distribution systems}.
\newblock \bibinfo{journal}{Journal of Water Resources Planning and Management} \bibinfo{volume}{148}, \bibinfo{pages}{04021106}.
\bibitem[{Sukhbaatar et~al.(2015)Sukhbaatar, Weston, Fergus et~al.}]{sukhbaatar2015end}
\bibinfo{author}{Sukhbaatar, S.}, \bibinfo{author}{Weston, J.}, \bibinfo{author}{Fergus, R.}, et~al., \bibinfo{year}{2015}.
\newblock \bibinfo{title}{End-to-end memory networks}.
\newblock \bibinfo{journal}{Advances in neural information processing systems} \bibinfo{volume}{28}.
\bibitem[{Truong et~al.(2023)Truong, Tello, Lazovik and Degeler}]{truong2023graph}
\bibinfo{author}{Truong, H.}, \bibinfo{author}{Tello, A.}, \bibinfo{author}{Lazovik, A.}, \bibinfo{author}{Degeler, V.}, \bibinfo{year}{2023}.
\newblock \bibinfo{title}{Graph neural networks for pressure estimation in water distribution systems}.
\newblock \bibinfo{journal}{arXiv preprint arXiv:2311.10579} .
\bibitem[{Veli{\v{c}}kovi{\'c} et~al.(2022)Veli{\v{c}}kovi{\'c}, Badia, Budden, Pascanu, Banino, Dashevskiy, Hadsell and Blundell}]{velivckovic2022clrs}
\bibinfo{author}{Veli{\v{c}}kovi{\'c}, P.}, \bibinfo{author}{Badia, A.P.}, \bibinfo{author}{Budden, D.}, \bibinfo{author}{Pascanu, R.}, \bibinfo{author}{Banino, A.}, \bibinfo{author}{Dashevskiy, M.}, \bibinfo{author}{Hadsell, R.}, \bibinfo{author}{Blundell, C.}, \bibinfo{year}{2022}.
\newblock \bibinfo{title}{The {CLRS} algorithmic reasoning benchmark}, in: \bibinfo{booktitle}{International Conference on Machine Learning}, \bibinfo{organization}{PMLR}. pp. \bibinfo{pages}{22084--22102}.
\bibitem[{Veli{\v{c}}kovi{\'c} and Blundell(2021)}]{velivckovic2021neural}
\bibinfo{author}{Veli{\v{c}}kovi{\'c}, P.}, \bibinfo{author}{Blundell, C.}, \bibinfo{year}{2021}.
\newblock \bibinfo{title}{Neural algorithmic reasoning}.
\newblock \bibinfo{journal}{Patterns} \bibinfo{volume}{2}.
\bibitem[{Veli{\v{c}}kovi{\'c} et~al.(2020a)Veli{\v{c}}kovi{\'c}, Buesing, Overlan, Pascanu, Vinyals and Blundell}]{velivckovic2020pointer}
\bibinfo{author}{Veli{\v{c}}kovi{\'c}, P.}, \bibinfo{author}{Buesing, L.}, \bibinfo{author}{Overlan, M.}, \bibinfo{author}{Pascanu, R.}, \bibinfo{author}{Vinyals, O.}, \bibinfo{author}{Blundell, C.}, \bibinfo{year}{2020}a.
\newblock \bibinfo{title}{Pointer graph networks}.
\newblock \bibinfo{journal}{Advances in Neural Information Processing Systems} \bibinfo{volume}{33}, \bibinfo{pages}{2232--2244}.
\bibitem[{Veli{\v{c}}kovi{\'c} et~al.(2018)Veli{\v{c}}kovi{\'c}, Cucurull, Casanova, Romero, Li{\`o} and Bengio}]{velivckovic2018graph}
\bibinfo{author}{Veli{\v{c}}kovi{\'c}, P.}, \bibinfo{author}{Cucurull, G.}, \bibinfo{author}{Casanova, A.}, \bibinfo{author}{Romero, A.}, \bibinfo{author}{Li{\`o}, P.}, \bibinfo{author}{Bengio, Y.}, \bibinfo{year}{2018}.
\newblock \bibinfo{title}{Graph attention networks}, in: \bibinfo{booktitle}{International Conference on Learning Representations}.
\bibitem[{Veli{\v{c}}kovi{\'c} et~al.(2020b)Veli{\v{c}}kovi{\'c}, Ying, Padovano, Hadsell and Blundell}]{velivckovic2020neural}
\bibinfo{author}{Veli{\v{c}}kovi{\'c}, P.}, \bibinfo{author}{Ying, R.}, \bibinfo{author}{Padovano, M.}, \bibinfo{author}{Hadsell, R.}, \bibinfo{author}{Blundell, C.}, \bibinfo{year}{2020}b.
\newblock \bibinfo{title}{Neural execution of graph algorithms}, in: \bibinfo{booktitle}{International Conference on Learning Representations}.
\bibitem[{Vrachimis et~al.(2020)Vrachimis, Eliades, Taormina, Ostfeld, Kapelan, Liu, Kyriakou, Pavlou, Qiu and Polycarpou}]{vrachimis2020battledim}
\bibinfo{author}{Vrachimis, S.G.}, \bibinfo{author}{Eliades, D.G.}, \bibinfo{author}{Taormina, R.}, \bibinfo{author}{Ostfeld, A.}, \bibinfo{author}{Kapelan, Z.}, \bibinfo{author}{Liu, S.}, \bibinfo{author}{Kyriakou, M.}, \bibinfo{author}{Pavlou, P.}, \bibinfo{author}{Qiu, M.}, \bibinfo{author}{Polycarpou, M.M.}, \bibinfo{year}{2020}.
\newblock \bibinfo{title}{Battledim: Battle of the leakage detection and isolation methods}, in: \bibinfo{booktitle}{Proc., 2nd Int. CCWI/WDSA Joint Conf}, pp. \bibinfo{pages}{1--6}.
\bibitem[{Wan et~al.(2022)Wan, Kuhanestani, Farmani and Keedwell}]{wan2022literature}
\bibinfo{author}{Wan, X.}, \bibinfo{author}{Kuhanestani, P.K.}, \bibinfo{author}{Farmani, R.}, \bibinfo{author}{Keedwell, E.}, \bibinfo{year}{2022}.
\newblock \bibinfo{title}{Literature review of data analytics for leak detection in water distribution networks: A focus on pressure and flow smart sensors}.
\newblock \bibinfo{journal}{Journal of Water Resources Planning and Management} \bibinfo{volume}{148}, \bibinfo{pages}{03122002}.
\bibitem[{Wang et~al.(2022)Wang, Lan, Liu, Ouyang, Qin, Lu, Chen, Zeng and Philip}]{wang2022generalizing}
\bibinfo{author}{Wang, J.}, \bibinfo{author}{Lan, C.}, \bibinfo{author}{Liu, C.}, \bibinfo{author}{Ouyang, Y.}, \bibinfo{author}{Qin, T.}, \bibinfo{author}{Lu, W.}, \bibinfo{author}{Chen, Y.}, \bibinfo{author}{Zeng, W.}, \bibinfo{author}{Philip, S.Y.}, \bibinfo{year}{2022}.
\newblock \bibinfo{title}{Generalizing to unseen domains: A survey on domain generalization}.
\newblock \bibinfo{journal}{IEEE transactions on knowledge and data engineering} \bibinfo{volume}{35}, \bibinfo{pages}{8052--8072}.
\bibitem[{Wu et~al.(2021)Wu, Pan, Chen, Long, Zhang and Philip}]{wu2021comprehensive}
\bibinfo{author}{Wu, Z.}, \bibinfo{author}{Pan, S.}, \bibinfo{author}{Chen, F.}, \bibinfo{author}{Long, G.}, \bibinfo{author}{Zhang, C.}, \bibinfo{author}{Philip, S.Y.}, \bibinfo{year}{2021}.
\newblock \bibinfo{title}{A comprehensive survey on graph neural networks}.
\newblock \bibinfo{journal}{IEEE transactions on neural networks and learning systems} \bibinfo{volume}{32}, \bibinfo{pages}{4--24}.
\bibitem[{Xhonneux et~al.(2021)Xhonneux, Deac, Veli{\v{c}}kovi{\'c} and Tang}]{xhonneux2021transfer}
\bibinfo{author}{Xhonneux, L.P.}, \bibinfo{author}{Deac, A.I.}, \bibinfo{author}{Veli{\v{c}}kovi{\'c}, P.}, \bibinfo{author}{Tang, J.}, \bibinfo{year}{2021}.
\newblock \bibinfo{title}{How to transfer algorithmic reasoning knowledge to learn new algorithms?}
\newblock \bibinfo{journal}{Advances in Neural Information Processing Systems} \bibinfo{volume}{34}, \bibinfo{pages}{19500--19512}.
\bibitem[{Xing and Sela(2022)}]{xing2022graph}
\bibinfo{author}{Xing, L.}, \bibinfo{author}{Sela, L.}, \bibinfo{year}{2022}.
\newblock \bibinfo{title}{Graph neural networks for state estimation in water distribution systems: Application of supervised and semisupervised learning}.
\newblock \bibinfo{journal}{Journal of Water Resources Planning and Management} \bibinfo{volume}{148}, \bibinfo{pages}{04022018}.
\bibitem[{Xu et~al.(2019)Xu, Hu, Leskovec and Jegelka}]{xu2019powerful}
\bibinfo{author}{Xu, K.}, \bibinfo{author}{Hu, W.}, \bibinfo{author}{Leskovec, J.}, \bibinfo{author}{Jegelka, S.}, \bibinfo{year}{2019}.
\newblock \bibinfo{title}{How powerful are graph neural networks?}, in: \bibinfo{booktitle}{International Conference on Learning Representations}.
\bibitem[{Xu et~al.(2020)Xu, Li, Zhang, Du, Kawarabayashi and Jegelka}]{xu2019can}
\bibinfo{author}{Xu, K.}, \bibinfo{author}{Li, J.}, \bibinfo{author}{Zhang, M.}, \bibinfo{author}{Du, S.S.}, \bibinfo{author}{Kawarabayashi, K.i.}, \bibinfo{author}{Jegelka, S.}, \bibinfo{year}{2020}.
\newblock \bibinfo{title}{What can neural networks reason about?}, in: \bibinfo{booktitle}{International Conference on Learning Representations}.
\bibitem[{Xu et~al.(2021)Xu, Zhang, Li, Du, Kawarabayashi and Jegelka}]{xu2020neural}
\bibinfo{author}{Xu, K.}, \bibinfo{author}{Zhang, M.}, \bibinfo{author}{Li, J.}, \bibinfo{author}{Du, S.S.}, \bibinfo{author}{Kawarabayashi, K.I.}, \bibinfo{author}{Jegelka, S.}, \bibinfo{year}{2021}.
\newblock \bibinfo{title}{How neural networks extrapolate: From feedforward to graph neural networks}, in: \bibinfo{booktitle}{International Conference on Learning Representations}.
\bibitem[{Zaheer et~al.(2017)Zaheer, Kottur, Ravanbakhsh, Poczos, Salakhutdinov and Smola}]{zaheer2017deep}
\bibinfo{author}{Zaheer, M.}, \bibinfo{author}{Kottur, S.}, \bibinfo{author}{Ravanbakhsh, S.}, \bibinfo{author}{Poczos, B.}, \bibinfo{author}{Salakhutdinov, R.R.}, \bibinfo{author}{Smola, A.J.}, \bibinfo{year}{2017}.
\newblock \bibinfo{title}{Deep sets}.
\newblock \bibinfo{journal}{Advances in neural information processing systems} \bibinfo{volume}{30}.
\bibitem[{Zanfei et~al.(2022)Zanfei, Brentan, Menapace, Righetti and Herrera}]{zanfei2022graph}
\bibinfo{author}{Zanfei, A.}, \bibinfo{author}{Brentan, B.M.}, \bibinfo{author}{Menapace, A.}, \bibinfo{author}{Righetti, M.}, \bibinfo{author}{Herrera, M.}, \bibinfo{year}{2022}.
\newblock \bibinfo{title}{Graph convolutional recurrent neural networks for water demand forecasting}.
\newblock \bibinfo{journal}{Water Resources Research} \bibinfo{volume}{58}, \bibinfo{pages}{e2022WR032299}.
\bibitem[{Zhou et~al.(2019)Zhou, Tang, Xu, Meng, Chu, Xin and Fu}]{zhou2019deep}
\bibinfo{author}{Zhou, X.}, \bibinfo{author}{Tang, Z.}, \bibinfo{author}{Xu, W.}, \bibinfo{author}{Meng, F.}, \bibinfo{author}{Chu, X.}, \bibinfo{author}{Xin, K.}, \bibinfo{author}{Fu, G.}, \bibinfo{year}{2019}.
\newblock \bibinfo{title}{Deep learning identifies accurate burst locations in water distribution networks}.
\newblock \bibinfo{journal}{Water research} \bibinfo{volume}{166}, \bibinfo{pages}{115058}.

\end{thebibliography}

\end{document}